\documentclass{article}

\usepackage{microtype}
\usepackage{graphicx}
\usepackage{caption}
\usepackage{subcaption}
\usepackage{booktabs} % for professional tables
\usepackage{soul}
\usepackage[mathscr]{eucal}
\usepackage{nicefrac}

\usepackage{hyperref}

\usepackage[accepted]{icml2023}

\usepackage{amsmath}
\usepackage{amssymb}
\usepackage{mathtools}
\usepackage{amsthm}
\usepackage{float}
\usepackage{wrapfig}
\usepackage{subcaption}
\usepackage{thmtools,thm-restate}

\def\given{\middle\vert}

\def\expectation{\mathbb{E}}

\def\defeq{\dot=}

\icmltitlerunning{Deep Laplacian-based Options for Temporally-Extended Exploration}

\begin{document}

\twocolumn[
\icmltitle{Deep Laplacian-based Options for Temporally-Extended Exploration}



\icmlsetsymbol{equal}{*}
\icmlsetsymbol{dagger}{$\dagger$}

\begin{icmlauthorlist}
\icmlauthor{Martin Klissarov}{mila,equal}
\icmlauthor{Marlos C. Machado}{dagger,amii,uofa,cifar}
\end{icmlauthorlist}

% \icmlaffiliation{dm_intern}{Work done during an internship at DeepMind}
\icmlaffiliation{mila}{Mila, McGill University}
%\icmlaffiliation{dm}{DeepMind}
\icmlaffiliation{amii}{Alberta Machine Intelligence Institute (Amii)}
\icmlaffiliation{uofa}{Department of Computing Science, University of Alberta} 
\icmlaffiliation{cifar}{Canada CIFAR AI Chair}

\icmlcorrespondingauthor{Martin Klissarov}{martin.klissarov@mail.mcgill.ca}


\vskip 0.3in
]



\printAffiliationsAndNotice{\icmlAdditionalNote\icmlSecondNote} 

\begin{abstract}
Selecting exploratory actions that generate a rich stream of experience for better learning is a fundamental challenge in reinforcement learning (RL). An approach to tackle this problem consists in selecting actions according to specific policies for an extended period of time, also known as options. A recent line of work to derive such exploratory options builds upon the eigenfunctions of the graph Laplacian. Importantly, until now these methods have been mostly limited to tabular domains where (1)~the graph Laplacian matrix was either given or could be fully estimated, (2)~performing eigendecomposition on this matrix was computationally tractable, and (3)~value functions could be learned exactly. Additionally, these methods required a separate option discovery phase. These assumptions are fundamentally not scalable. In this paper we address these limitations and show how recent results for directly approximating the eigenfunctions of the Laplacian can be leveraged to truly scale up options-based exploration. To do so, we introduce a fully online deep RL algorithm for discovering Laplacian-based options and evaluate our approach on a variety of pixel-based tasks. We compare to several state-of-the-art exploration methods and show that our approach is effective, general, and especially promising in non-stationary settings.

\end{abstract}

\section{Introduction}
In reinforcement learning (RL), an agent interacts with an unknown environment in order to maximize the sum of rewards. At each step of this interaction the agent receives an observation, a reward signal, and it takes an action. Importantly, the actions the agent takes do not only impact the reward it receives but also its stream of experience, that is, its future observations and affordances. Thus, there are two types of actions the agent can take: actions that exploit what it has learned so far and actions, known as exploratory actions, that are seemingly suboptimal but that can lead to additional information about the environment. A fundamental challenge in RL resides in selecting exploratory actions that generate a rich stream of experience for better learning. 

In this paper we build on the idea of promoting exploration by explicitly considering courses of actions within the environment. That is, the agent explores by selecting actions according to a  specific (not random) policy for an extended period of time. This temporally-extended exploration allows the agent to be purposeful and to ensure that parts of the environment that are unlikely to be visited by chance can be visited more often~\citep{machado16learning,jinnai2019discovering,ecoffet2021first}. We model temporally-extended exploration through options~\citep{precup00temporal,sutton1999between}, a well-known formalism for representing behaviour at different timescales. 
Methods based on these ideas have been quite successful: domains of application include Atari 2600 games~\citep{ecoffet2021first,dabney2021temporally}, continuous control on the MuJoCo simulator \citep{park202lipschitz}, and real-world problems such as balloon navigation~\citep{bellemare2020autonomous}. \looseness=-1

A central question around temporally-extended exploration methods is how one should define such temporal abstractions. Put differently, \emph{how should one discover options for exploration?} Solutions often revolve around extremely simple approaches such as action repetition~\citep[e.g.,][]{dabney2021temporally} or approaches that exploit domain-specific information~\citep[e.g.,][]{bellemare2020autonomous,ecoffet2021first}. Potentially more general methods for temporally-extended exploration have not been convincingly shown to truly scale beyond small domains~\citep[e.g.,][]{machado2017laplacian,machado2018eigenoption,jinnai2019discovering,jinnai2020exploration,bar2020option}; we do so here.

Specifically, we consider the line of work that discovers exploratory options by leveraging a diffusion model that encodes the flow of information in the environment's underlying graph~\citep[c.f.][]{machado21temporal}. The diffusion model of choice is the graph Laplacian \citep{chung1997spectral} as its eigenfunctions are known to capture the topology of the environment at various timescales \citep{mahadevan2007proto}. The studied methods ground the option discovery process in the eigenfunctions of the Laplacian to encourage the agent to follow the principal directions of the state space in an attempt to obtain effective temporally-extended exploration.\looseness=-1

Importantly, as aforementioned, some of the most compelling empirical results around Laplacian-based methods are limited to tabular domains. It is uncommon to see similar gains with deep function approximation~\citep{machado2018eigenoption,jinnai2020exploration}. This performance gap could be attributed to several factors, including the need to perform an eigendecomposition on the $|\mathscr{S}| \times |\mathscr{S}|$ graph Laplacian matrix, where $|\mathscr{S}|$ is the number of states in the environment. In large domains, obtaining this matrix is difficult and the cost of performing its eigendecomposition is prohibitive. Additionally, these tabular methods often rely on strong assumptions, such as access to the full eigenspectrum of the graph Laplacian matrix and, in order to define options, access to accurate value estimates. Naturally, these things are impossible under function approximation. Finally, most methods rely on multiple well-defined stages, such as pre-training a full set of Laplacian-based options only to later maximize the environment reward~\citep{bar2020option,machado21temporal}. This adds extra complexity and can be impractical when considering real-world domains.

In this paper, we extend, in a general way, the Laplacian-based options framework to deep function approximation, presenting solutions to all the aforementioned difficulties.  Specifically, we (1) validate, for option discovery, the feasibility of using objectives that approximate the eigenfunctions of the Laplacian with neural networks \citep{wu2019laplacian,wang2021towards}. This does not only allow us to circumvent the cubic cost of performing eigendecomposition but it gives us an anytime approach for approximating these eigenfunctions. We (2) introduce a fully online algorithm for discovering Laplacian-based options that promote exploration. This algorithm, termed \textit{deep covering eigenoptions}, also gracefully deals with conditions for option termination and value function approximation. We validate the proposed approach by (3) benchmarking, for the first time, Laplacian-based methods beyond navigation tasks against several state-of-the-art exploration methods such as count-based \citep{bellemare2016unifying}, diversity-based \citep{eysenbach2019diversity} and prediction-based methods \citep{burda2019exploration}. Finally, we (4) illustrate the effectiveness of Laplacian-based option-driven exploration when faced with non-stationary environments, opening the way for continual exploration. \looseness=-1

\section{Preliminaries}\label{sec:background}

We consider an agent interacting with an environment in which at time step $t$ the agent is in state $S_t \in \mathscr{S}$, selects an action $A_t \in \mathscr{A}$, and in response the environment emits a scalar reward $R_{t+1}$ and transitions to a new state, $S_{t+1} \in \mathscr{S}$, according to a transition probability kernel $p(s'|s,a) = \text{Pr}(S_{t+1} = s' | S_t = s, A_t = a)$. The agent's goal is to find a policy $\pi: \mathscr{S} \to \Delta(\mathscr{A})$ that maximizes the expected discounted sum of rewards, $\mathbb{E}_{\pi}\left[ \sum_{i=t}^\infty \gamma^{i-t} R_{i+1} \right] \equiv \mathbb{E}_{\pi}\left[G_t \right]$, where $\gamma \in [0,1)$ is the discount factor. \looseness=-1

We focus on value-based methods that obtain the policy $\pi$ by estimating the state-action value function $q_\pi(s,a) = \mathbb{E}_\pi\left[G_t \given S_t = s, A_t = a \right]$. Q-learning~\citep{watkins1992q} is likely the most used algorithm to estimate the optimal policy, $\pi^*$. In the tabular case, where we assign a value to each state, its update rule is 
$Q (S_t, A_t) \gets Q (S_t, A_t) + \alpha \left[ R_{t+1} + \gamma \max_{a \in \mathscr{A}} Q(S_{t+1}, a) - Q(S_t, A_t) \right]$. 

As the complexity of the tasks we tackle increases, we need to approximate the optimal value function through a parameterized function $Q_{\boldsymbol{\theta}_t}(s,a)$. We use neural networks for parameterizing $Q$, as in the Deep Q-Networks (DQN) algorithm \citep{mnih2013playing,mnih2015human}. An important property of DQN is that it is off-policy, that is, it does not rely on data generated by the current policy $\pi$ to improve its performance. Instead it uses a replay buffer \citep{lin1992self} of stored past experience from which it samples batches of transitions. 

In this paper we use the Double DQN algorithm~\citep{hasselt2016deep} with $n$-step targets~\citep{sutton2018reinforcement,hessel2018rainbow}, which are improvements over DQN. Double DQN uses an alternative double estimator method to address the issue DQN has of overestimating the action values, and $n$-step targets consist in explicitly accumulating rewards over multiple steps instead of bootstrapping from the action value at time $t+1$. The update rule we use after integrating both Double DQN and $n$-step targets is
\begin{align}
\boldsymbol{\theta}_{t+1} \gets \boldsymbol{\theta}_t + \alpha \left[ Y_{t}^{(n)} - Q_{\boldsymbol{\theta}_t}(S_t, A_t)\right] \nabla_{\boldsymbol{\theta}_t}  Q_{\boldsymbol{\theta}_t}(S_t, A_t)
% \nonumber
\label{ddqn}
\end{align}
\vspace{-5mm}
\begin{align}
Y_{t}^{(n)} = R_{t+1}^{(n)} +
\gamma^n Q_{\boldsymbol{\theta}^-_t}(S_{t+n}, \arg\max_{a' \in \mathscr{A}} Q_{\boldsymbol{\theta}_t}(S_{t+n}, a')),
\nonumber
% \label{ddqn}
\end{align}
where $R_{t+1}^{(n)} \defeq{} \sum_{i=1}^{n-1} \gamma^i R_{t+i+1}$, and $\boldsymbol{\theta}^-_t$ denotes the parameters of a duplicate network, which are updated less often for stability purposes. 

\textbf{Options. } The agent-environment interaction occurs at discrete low-level timesteps, which we index by $t$. However, an intelligent agent will pursue a variety of goals and reason over a hierarchy of timescales. The options framework \citep{sutton1999between} provides a formalism for such abstractions over time and is one of the most common frameworks in hierarchical reinforcement learning (HRL). An option within the option set $\mathscr{O}$ is composed of an intra-option policy $\pi(a|s,o)$ to select actions, a termination function $\beta(s,o)$ to determine when an option stops executing, and an initiation set $\mathscr{I}(s,o)$ to restrict the states in which an option may be selected. We may also choose which option to execute in a specific state through the policy over options $\pi_{\mathscr{O}}: \mathscr{S} \to \Delta(\mathscr{O})$.  The options' policies are often learned through an intrinsic reward function $r^o(s, a, s')$. % 

\section{Covering Eigenoptions}\label{sec:ceo}

Although the options framework provides a useful formalism for temporal abstraction, it does not specify how we may obtain useful abstractions. Indeed, option discovery remains a fundamental challenge in HRL for which a wide variety of strategies exists, whether it be through feudal RL
\citep{dayan1992feudal, vezhnevets2017feudal}, 
directly maximizing the environment's reward with policy-gradients \citep{bacon2017option,harb2018when}, maximizing mutual-information-based objectives \citep{gregor16variational,eysenbach2019diversity, kim2021unsupervised}, skill chaining \citep{konidaris2009skill, bagaria2020option}, or probabilistic inference \citep{smith2018inference,wulfmeier2021data}. In this paper we leverage a particular class of methods that ground the option discovery process in a learned representation of the environment, the eigenfunctions of the graph Laplacian. Specifically, we focus on the recently introduced \textit{covering eigenoptions}~\cite{machado21temporal}. \looseness=-1

Covering eigenoptions (CEO) are obtained by an iterative process called the representation-driven option discovery (ROD) cycle. The cycle has 3 mains steps: 1.~The agent collects samples from the environment and uses them to learn a representation. 2.~This representation is used to define intrinsic rewards, which are used to learn the options. 3.~The options empower the agent such that, when back to the first step, it can behave in ways it was not likely to behave before. As a consequence, it can refine and improve its representations from which new options will be derived and later executed, continuing in a never-ending virtuous cycle. \looseness=-1

CEO instantiates the ROD cycle by using as representation the eigenfunctions of the graph Laplacian. The graph Laplacian is a diffusion model that encodes the way information flows on a graph. In RL, the Laplacian encodes the environment's underlying graph, where nodes represent states and edges encode transitions between such states. Of particular interest are the eigenfunctions of the Laplacian, which capture the dynamics of the environment at different timescales ~\citep{mahadevan2005proto}. A set of eigenfunctions can then be used to obtain a rich representation of the environment, to which we refer to as the \textit{Laplacian representation}. CEO defines option $o_i$ as the policy that maximizes the intrinsic reward defined by the eigenfunction $\textbf{e}_i$ of the Laplacian:
$$r^{\textbf{e}_i}(s,s') = \textbf{e}_i(s') - \textbf{e}_i(s).$$
The option terminates in every state $s$ where $q_\pi^{\textbf{e}_i}(s,a) \leq 0$ for all $a \in \mathscr{A}$, where $q_\pi^{\textbf{e}_i}$ is defined w.r.t. $r^{\textbf{e}_i}(\cdot, \cdot)$.

The empirical results pertaining CEO show that it is more efficient at covering the classic Four rooms domain \citep{sutton1999between} by more than an order of magnitude when compared to a random walk. However, that is the extent in which CEO has been empirically evaluated. It is not clear, for example, whether CEO's strong performance can generalize to different environments, or whether it can be leveraged within a reward maximization algorithm. This validation is important before we try to scale up such an algorithm. \looseness=-1

\begin{figure}[t]
     \centering
     \begin{subfigure}[b]{0.3\columnwidth}
         \centering
         \includegraphics[width=\columnwidth]{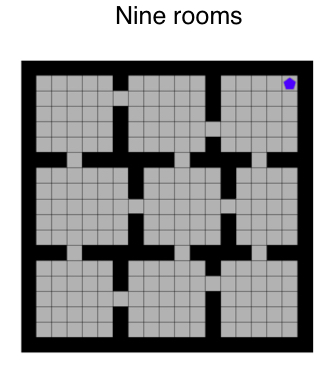}
     \end{subfigure}
     \hfill
     \begin{subfigure}[b]{0.3\columnwidth}
         \centering
         \includegraphics[width=\columnwidth]{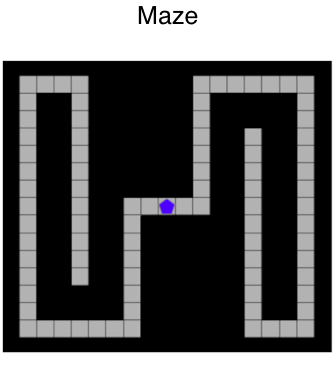}
     \end{subfigure}
     \hfill
     \begin{subfigure}[b]{0.3\columnwidth}
         \centering
         \includegraphics[width=\columnwidth]{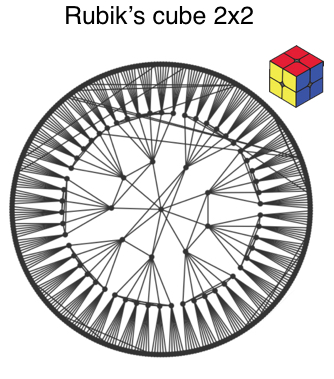}
     \end{subfigure}
        \caption{\textbf{Environments}: Nine rooms, Maze, and a subset of states in Rubik's cube 2x2. Their underlying topologies are quite different and they present different challenges in terms exploration.  All environments are stochastic: the agent's actions are randomly overwritten with probability $0.15$.}
        \label{fig:environments}
\end{figure}

We provide answers to the questions above in Appendix~\ref{sec:results_tabular}. We evaluate CEO in environments with different topologies (c.f. Figure~\ref{fig:environments} and detailed description in Appendix~\ref{app:environment_description}) while comparing its performance against well-established exploration algorithms. We also extend CEO beyond state coverage, obtaining an efficient reward maximizing method. We show, across environments with different topologies, that CEO is extremely effective in covering the state space, that it does so in a much more purposeful way, and that it does lead to faster reward maximization, even when considering the initial cost of option discovery. Besides a \textit{random policy}, we used \textit{count-based exploration} and $\epsilon$\textit{z-greedy}~\citep{dabney2021temporally} as baselines.

\section{Approximate Laplacian-based Options}

As aforementioned, a fundamental limitation of CEO is that, in order to define the intrinsic rewards used in the option discovery process, it relies on performing a costly eigendecomposition operation to obtain the eigenfunctions of the graph Laplacian. Estimating the full Laplacian matrix  in environments with large state spaces is impractical and, even if such a matrix was available, performing its eigendecomposition would not be scalable due to its cubic cost. This is a problem Laplacian-based option discovery methods need to face in order to be broadly applicable. In this section we show how recent methods that use neural networks to approximate the eigenfunctions of the Laplacian \citep[e.g.,][]{wu2019laplacian,wang2021towards} can be used in this framework.\looseness=-1

\begin{figure*}[t]
     \centering
     \begin{subfigure}[b]{0.3\textwidth}
         \centering
         \includegraphics[width=\textwidth]{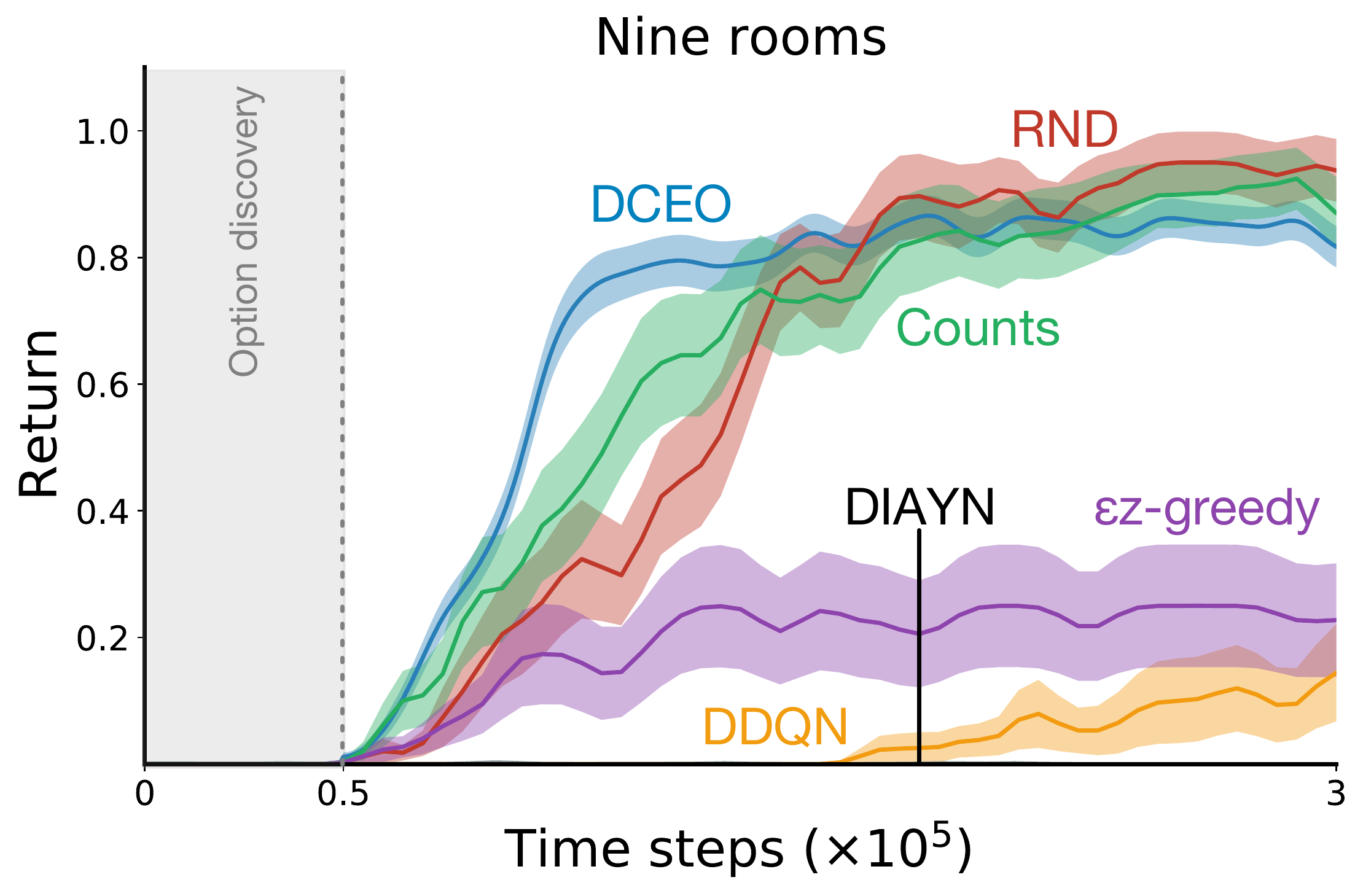}
     \end{subfigure}
     \hfill
     \begin{subfigure}[b]{0.3\textwidth}
         \centering
         \includegraphics[width=\textwidth]{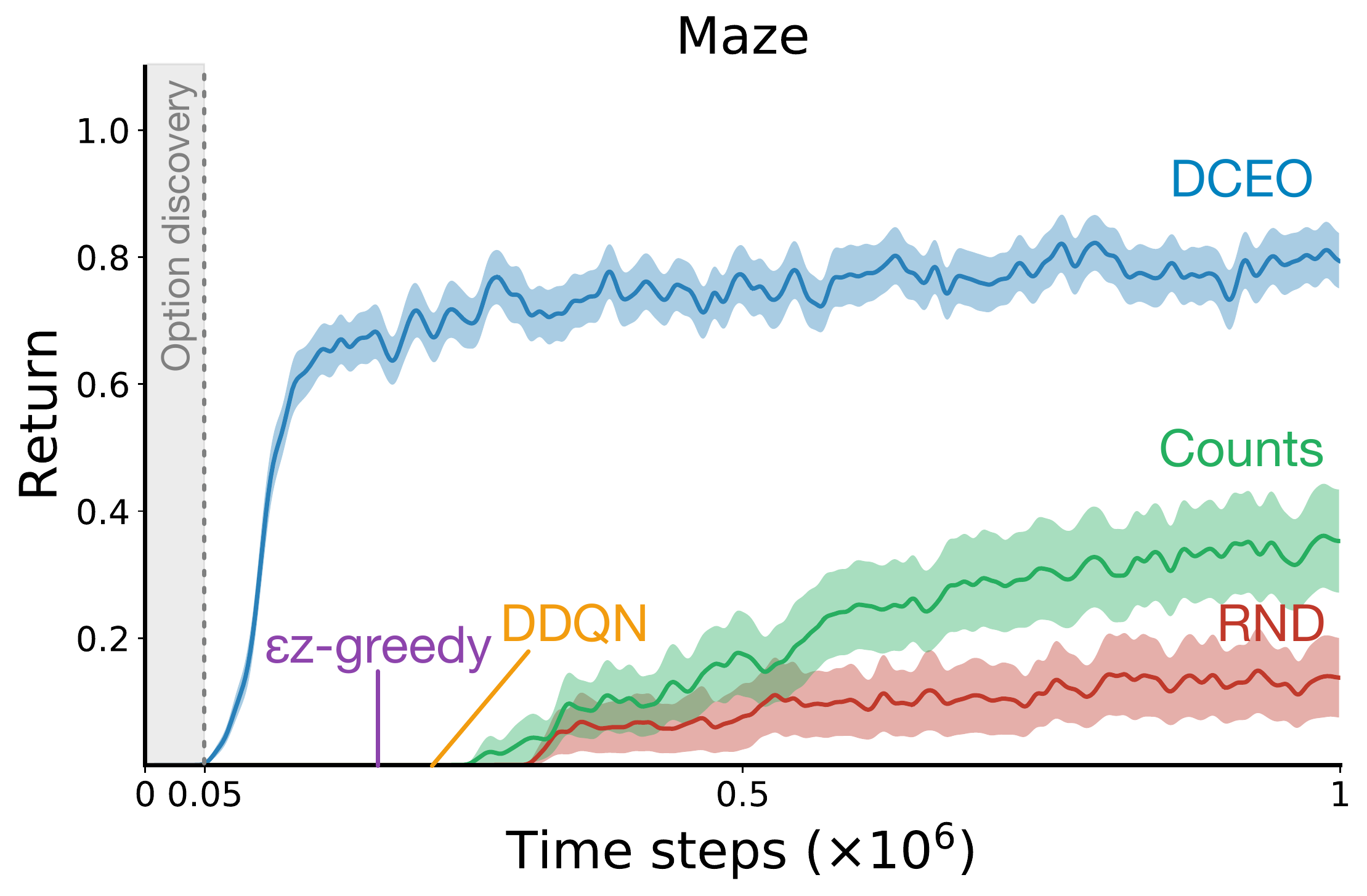}
     \end{subfigure}
     \hfill
     \begin{subfigure}[b]{0.3\textwidth}
         \centering
         \includegraphics[width=\textwidth]{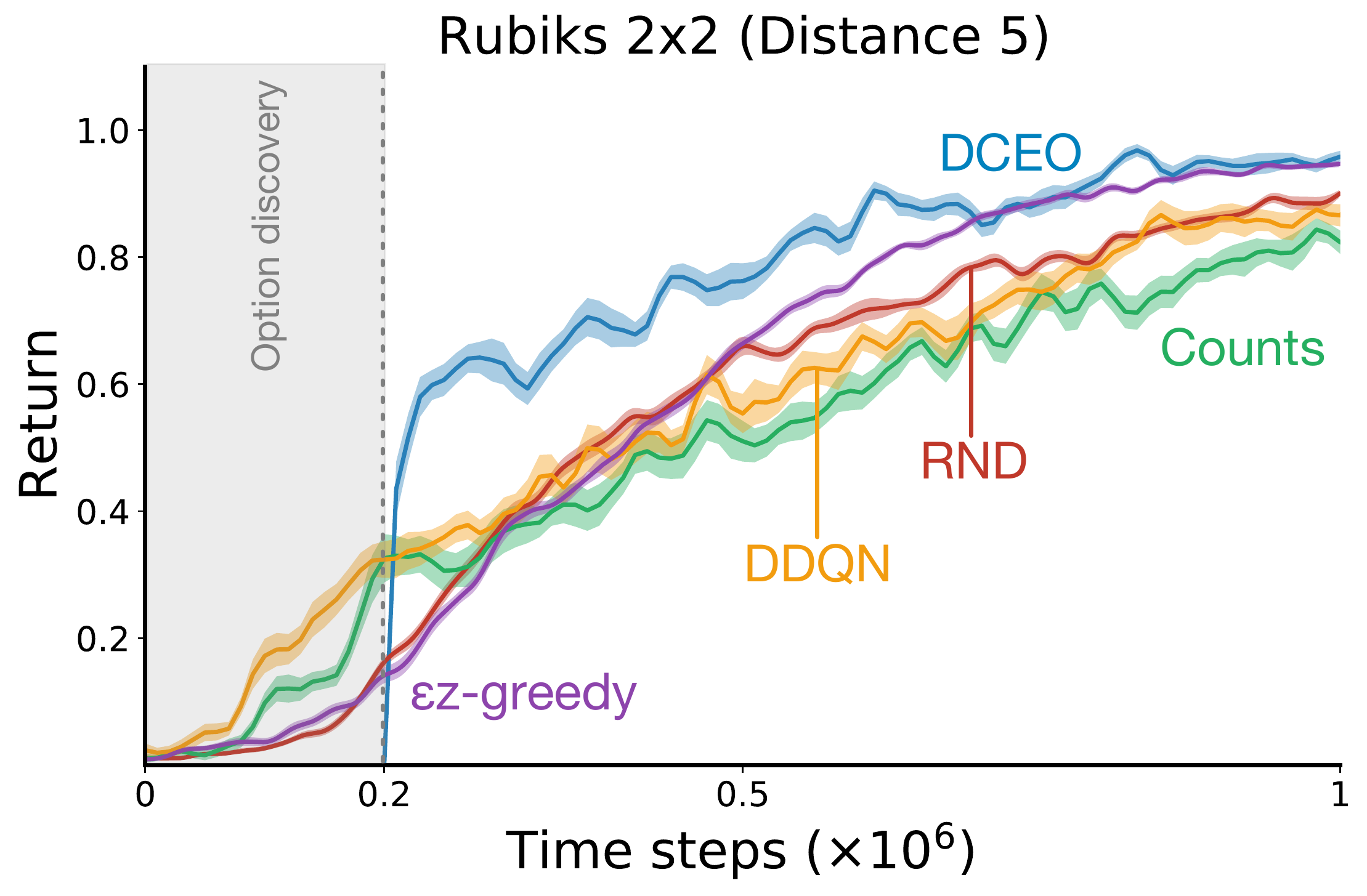}
     \end{subfigure}
        \caption{\textbf{Reward maximization}. DCEO uses a two-phased algorithm where it first pretrains a set of options through intrinsic motivation before leveraging them for reward maximization. DCEO's curve is delayed by the amount of time spent in the first phase of option discovery. Despite this additional cost, we notice that it produces strong performance across all domains. When the curves of different methods are not visible inside the shaded region it is because the exploration strategy used by the baseline method did not lead to a single positive reward during the whole period. Results show the mean and standard deviation across 30 seeds.}
        \label{fig:fa_return_maximization}
\end{figure*}

Specifically, we use a direct approximation of the  eigenfunctions\footnote{Eigenfunctions can be understood as a generalization of eigenvectors that allow for the same objective to be derived for the case of a continuous state space. See derivation by \citet{wu2019laplacian}.} of the Laplacian through an objective borrowed from graph theory~\citep{koren2005drawing}. It was first proposed by \citet{wu2019laplacian} to be adapted to the RL setting, and recently extended by \citet{wang2021towards}. We consider the objective
\begin{align*}
    \min_{\boldsymbol{f}_1, ... \boldsymbol{f}_d} \quad & \sum_{i=1}^d c_i \boldsymbol{f}_i^{\top} L \boldsymbol{f}_i 
    & \textrm{s.t.} \quad & \boldsymbol{f}_i^{\top} \boldsymbol{f}_j = \delta_{ij} \forall i, j 
\end{align*}

where $\{ \boldsymbol{f}_i \}_{i=1}^d$ are approximations to the $d$ smallest eigenfunctions of the Laplacian, $c_i$ are their associated coefficients, and $\delta_{ij}$ is the delta Dirac that is 1 only when $i=j$ and it is 0 otherwise. We refer to this objective as the \textit{generalized Laplacian}. If the coefficients $c_i$ are chosen to be strictly decreasing, \citet{wang2021towards} has shown that its optimal solution are the eigenfunctions of the graph Laplacian.

To make this objective amenable to online RL, we define a loss function by rewriting this objective as an expectation:
\begin{align}
    G(f_1, ..., f_d) = \frac{1}{2} \expectation_{\pi} \left[ \sum_{i=1}^d \sum_{k=1}^i \big(f_k(s) - f_k(s')\big)^2 \right] + \nonumber \\ \beta \sum_{i=1}^d \sum_{j=1}^i \sum_{k=1}^i \Big( \expectation_{\pi}\big([f_j(s) f_k(s) - \delta_{jk}) \big] \Big)^2,
\label{gl}
\end{align}
where $\beta$ is the Lagrange multiplier and the coefficients are defined as $c_i = d - i + 1$. Intuitively, the first part ensures smoothness while the second incentivizes orthogonality between eigenfunctions.

Importantly, \citet{wang2021towards} has already used the generalized Laplacian to discover options, evaluating the agent's capacity to navigate between rooms in a gridworld. Nevertheless, besides the small scale of the experiments, they were conducted with random restarts throughout the environment, providing the agent, at no cost, with a complete picture of the task, side-stepping the problem of exploration. Indeed, recent results suggest that random restarts are quite important to guarantee the good performance reported by methods that use graph drawing objectives to approximate the eigenfunctions of the Laplacian \citep{erraqabi2021temporal}. Moreover, the inputs to the neural network optimizing the generalized Laplacian consisted of the agent's $(x,y)$ coordinates, which significantly simplifies the problem.

Thus, until now it has been an open question whether the generalized Laplacian objective could be leveraged within an algorithm that iteratively seeks to discover unknown parts of the state space; or whether it was effective with different types of inputs (e.g., pixels). Importantly, the discovered options were also never evaluated in their ability to improve the agent's capacity to maximize reward. In the next section we introduce an algorithm that extends CEO to deep function approximation, and we present empirical results showing how it addresses the limitations aforementioned.\looseness=-1

\begin{figure*}[t]
    \centering
    \includegraphics[width=.95\textwidth]{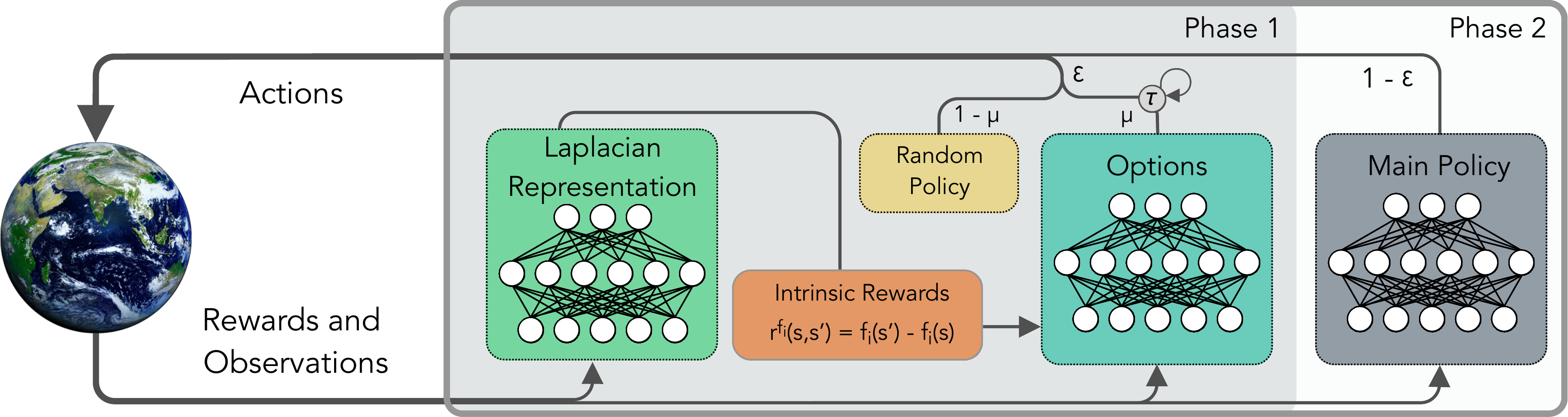}
    \caption{\textbf{Deep Covering Eigenoptions} (DCEO) algorithm. The agent receives rewards an observations and leverages them to learn the Laplacian representation which encodes the environment's topology at different timescales~\citep{mahadevan2005proto,mahadevan2007proto}. Using this representation the agent derives a set of intrinsic rewards to learn exploratoy options. It then selects actions either by being greedy w.p. $1-\epsilon$, or by exploring w.p. $\epsilon$. When taking exploratory actions these may come either from a random policy w.p. $1-\mu$, or from the set of options w.p. $\mu$. Once selected, the agent acts according to the option's policy until termination, which we denote by $\tau$. }
    \label{fig:algo_fig}
\end{figure*}

\section{A Two-Phased Scalable Method}

We first ask whether the generalized Laplacian is conducive to be used by option discovery methods. This is not necessarily trivial because Laplacian-based options are mostly defined by the intrinsic rewards generated by the eigenfunctions of the Laplacian and it is not clear whether the unavoidable approximation errors lead to bad reward functions. Moreover, when also using deep function approximation for learning both the main policy and the options' policies it is not clear whether these different sources of approximation can render learning unfeasible. Finally, how should one deal with option termination? As previously mentioned, Laplacian-based option discovery methods rely on very accurate value estimates to define termination, which can be impossible with function approximation.

In this section we introduce \emph{deep covering eigenoptions (DCEO)}, a two-phased algorithm that extends CEO to use deep function approximation in all of its steps. We validate its efficacy on pixel-based versions of the environments in Figure \ref{fig:environments}. Our results demonstrate that the generalized Laplacian is indeed conducive to be used by option discovery methods, at least when also considering the algorithmic choices we propose (c.f. Figure~\ref{fig:fa_return_maximization}). In fact, DCEO often outperforms several other baselines considered to be state-of-the-art methods. Below we properly introduce DCEO before going into details about the empirical methodology.
\vspace{-4pt}
\paragraph{Deep Covering Eigenoptions.} DCEO has two phases: the agent first interacts with the environment while learning the Laplacian representation and its corresponding set of options for $T_{discovery}$ steps. These options are then fixed and used by the agent to explore and maximize return.

Specifically, in the discovery phase, the agent learns options by maximizing intrinsic rewards based on the approximated eigenfunctions of the graph Laplacian.  When learning option $o_i$, following transition $s$ to $s'$, the agent is rewarded with
$r^{f_i}(s, s') = f_i(s') - f_i(s)$,
where $f$ is obtained from optimizing Equation \ref{gl}, with $f_i$ denoting the $i$th eigenfunction according to the order induced by the eigenvalues. For the termination function, instead of relying on perfectly accurate value estimates, we define option termination to be uniformly random with probably $\nicefrac{1}{D}$, where $D$ is the expected option length. We use $D = 10$ in all experiments.

For the reward maximization phase, the DCEO agent learns to maximize reward with DDQN and $n$-step targets (c.f. Eq.~\ref{ddqn}), and it uses $\epsilon$-greedy as the exploration strategy. The discovered options have an impact when the agent takes an exploratory step: with probability $\mu$ the agent does not take only a simple random primitive action, but it instead acts according to a sampled option's policy until it terminates, denoted by $\tau$, thus exploring in a temporally-extended way.

Figure~\ref{fig:algo_fig} presents an overview of DCEO and Algorithm~\ref{algo:two_phase_dceo_option_discovery}, in Appendix~\ref{app:algo}, is a more precise description of DCEO.
\vspace{-4pt}
\paragraph{Evaluation Procedure.} As aforementioned, we validate DCEO's efficacy on pixel-based versions of the environments in Figure~\ref{fig:environments}, with episodes being at most 100 steps long. In the navigation tasks the agent is given pixel images of the grid which are processed through a three-layered convolutional network. For the Rubik's cube, the agent observes the usual sticker colors, which are concatenated in a vector~\cite{agostinelli2019solving}. The agent processes this input with a three-layered fully connected network.

Besides a \emph{random policy}, which we label DDQN, we use \emph{count-based exploration}, $\epsilon$\emph{z-greedy}~\citep{dabney2021temporally}, \emph{RND}~\citep{burda2019exploration}, and \emph{DIAYN}~\citep{eysenbach2019diversity} as baselines. Count-based exploration consists in providing the agent with an intrinsic reward of $\nicefrac{1}{\sqrt{n(s)}}$ at each step, where $n(s)$ is the number of times the agent has visited state $s$ --- in \textbf{all} experiments we assume the agent has access to perfect counts, in fact making this an unfair baseline. 
We consider $\epsilon$z-greedy because it is an option-based exploration algorithm that uses action repetition to obtain temporally-extended exploration, and it has shown significant performance improvements on Atari 2600 games \citep{bellemare2013arcade,machado2018revisiting}. We consider DIAYN as another representative method that uses options for exploration, but based on a mutual information objective. Finally, we also consider RND, an error prediction-based exploration method with state-of-the-art performance across many environments. For the experiments on reward maximization, all methods are implemented on top of an n-step Double DQN (DDQN) baseline with $n=5$. Details on parameter tuning for each method, and the parameters used, are available in Appendix~\ref{app:experimental_details}.

Importantly, we implemented DCEO such that the networks used to learn options and to learn the Laplacian representation do not share weights with the network maximizing the environment rewards. This design choice ensures the auxiliary task effect~\citep{sutton2011horde,lyle2021effect} does not benefit DCEO. Therefore, DCEO's improvements in performance may only come from leveraging a set of Laplacian-based options that generate a rich and diverse stream of experience for learning.

\paragraph{Empirical Analysis} We evaluate DCEO in terms of its effectiveness in covering the state space and in empowering agents to learn to maximize the environment reward faster.\looseness=-1

For \emph{state coverage}, we 
considered DCEO's performance during the pretraining phase. The results depicted in Figure~\ref{fig:fa_coverage_maximization} in Appendix \ref{app:coverage_fa} show that DCEO is either significantly better or on-par to the other baselines. These results are the first evidence to indicate that approximating the eigenfunctions of the Laplacian through the generalized Laplacian objective is an effective approach, even without random restarts and when using high dimensional inputs.

We present the results for \emph{reward maximization} in Figure~\ref{fig:fa_return_maximization}. \footnote{The results on the \textit{Rubik's cube} were obtained with a tabular actor as the DDQN agent was not able to learn with any exploration method, whether it was DCEO or not. The generalized Laplacian was still optimized with neural networks, attesting to its robustness.} Overall, DCEO performs at least as well as well-established baselines such as Counts and RND. The performance of DIAYN is only shown for the \textit{Nine rooms} domain as it was not able to improve upon the DDQN baseline despite having access to the same amount of pretraining as DCEO. The difference in performance between DCEO and all baselines is the greatest on the \textit{Maze} experiment, which highlights the importance of temporally-extended exploration. More than outperforming baseline methods, these results are important because they establish Laplacian-based options as viable and competitive solutions for exploration problems.

\vspace{-0.2cm}
\section{A Single Continuous Cycle}

The results in the previous section provide the first strong evidence as to the effectiveness of Laplacian-based options for exploration under deep function approximation. However, the performance crucially relies on an initial phase dedicated to option discovery. This can be problematic as it implicitly assumes there are special phases in the learning period and that the world will not change after this phase, precluding further learning and reducing the agent's ability to adapt to the latest stream of experience; not to mention the introduction of additional hyperparameters. In this section, we go beyond these limitations and introduce a fully online and generally applicable algorithm. 

\vspace{-0.1cm}
\subsection{Online Discovery of Laplacian-based Options}

\begin{figure}[b]
     \centering
     \begin{subfigure}[b]{0.48\columnwidth}
         \centering
         \includegraphics[width=\columnwidth]{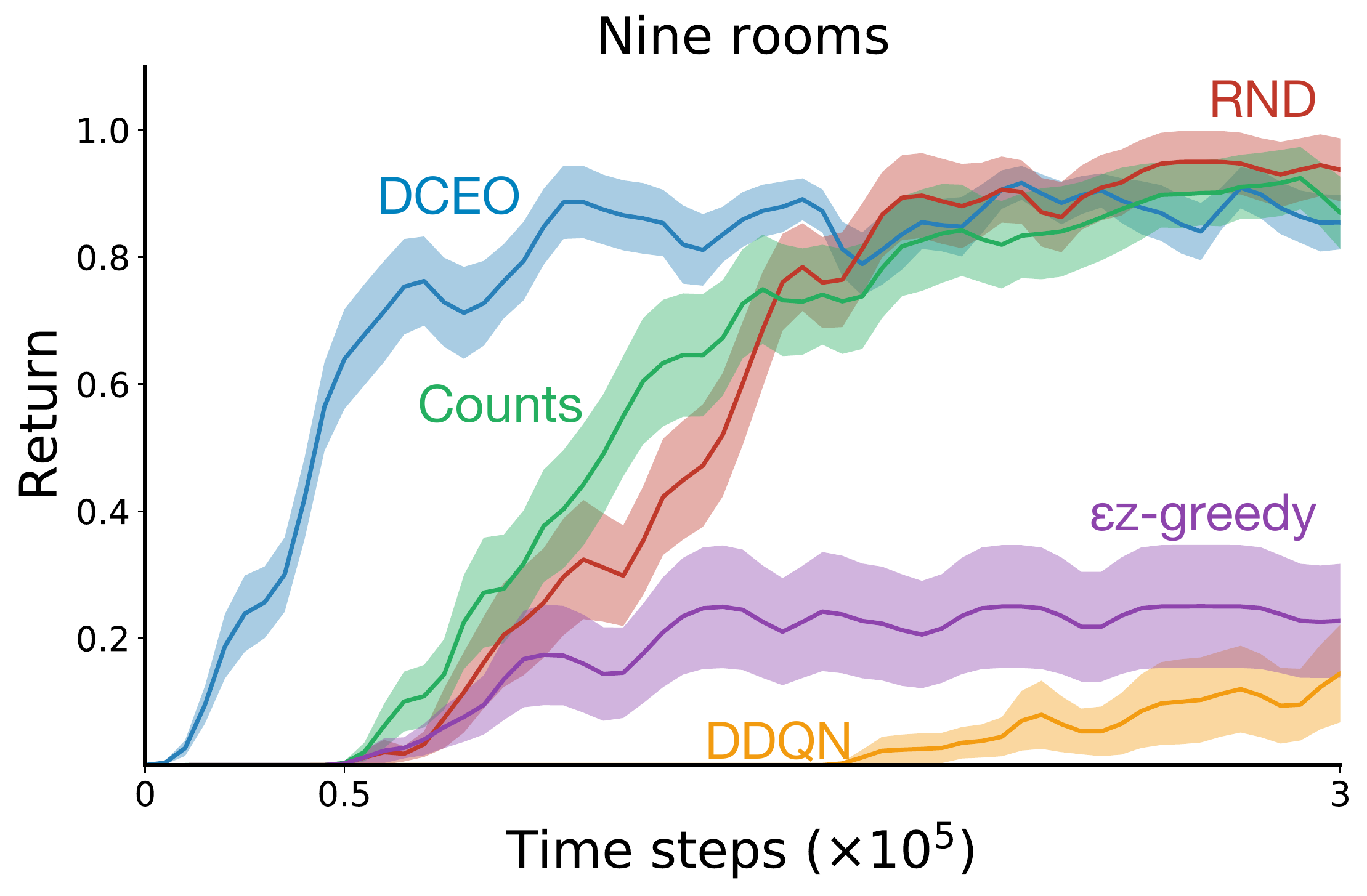}
     \end{subfigure}
     ~
     \begin{subfigure}[b]{0.48\columnwidth}
         \centering
         \includegraphics[width=\columnwidth]{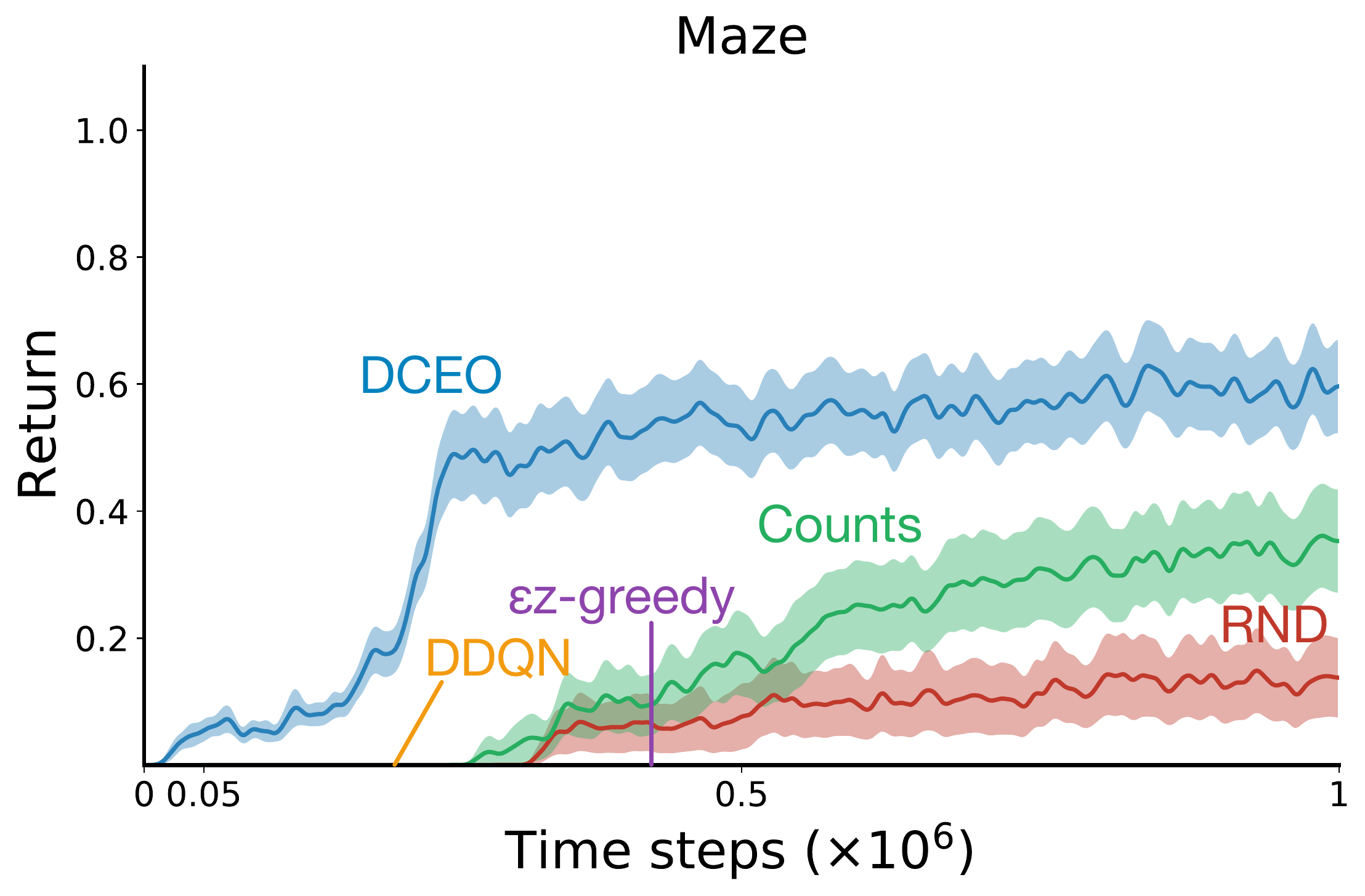}
     \end{subfigure}
        \caption{\textbf{Return maximization} using a fully online DCEO algorithm. Starting from a randomly initialized option set and Laplacian representaion, it learns to maximize reward as effectively as the two-phased algorithm. Results show the mean and standard deviation across 30 seeds.}
        \label{fig:fa_return_maximization_online}
\end{figure}

\begin{algorithm}[th!]
\small
\caption{Fully Online DCEO Algorithm}
\begin{algorithmic}[1]
\FOR{$i = 1$ {\bfseries to} $T$}
\IF{$i==1 \vee \bot$}
\STATE $\tau \leftarrow \textrm{True}$ \# \textit{Option termination}
\STATE $o \leftarrow -1$ \# \textit{No active option}
\STATE Reset environment and observe state $s$
\ENDIF
\STATE $\tau \leftarrow U(0,1) < \nicefrac{1}{D} \text{ } \vee  \text{ } \tau$
\IF{$\tau$}
\IF{$U(0, 1) < \epsilon$}
\IF{$U(0,1) < \mu$}
\STATE $o \leftarrow U(\mathscr{O})$; $\tau \leftarrow \textrm{False}$; $a \sim \pi_o(\cdot|s)$
\ELSE
\STATE $o \leftarrow -1$; $\tau \leftarrow \textrm{True}$; $a \sim U(\mathscr{A})$
\ENDIF
\ELSE
\STATE $a \leftarrow \max_{a \in \mathscr{A}} Q(s,a)$
\ENDIF
\ELSE
\STATE $a \sim \pi_o(\cdot|s)$
\ENDIF
\STATE Execute $a$, observe $r, s', \bot$ \# \textit{(}$\bot$\textit{ is episode termination)}
\STATE Store transition $(s,a,r,s')$ in buffer $B$; $s \leftarrow s'$
\STATE Sample a minibatch of transitions $(s_j, a_j, r_{j+1}, s_{j+1})$
% \STATE $\forall o \in \mathscr{O}, r_j \leftarrow f_o(s') - f_o(s)$
\STATE Train each option with intrinsic reward (Eq. \ref{ddqn})
\STATE Minimize the generalized Laplacian (Eq. \ref{gl})
\STATE Train main learner on extrinsic reward (Eq. \ref{ddqn})
\ENDFOR
\end{algorithmic}
\label{algo:dceo_online}
\end{algorithm}

We preserve the general structure of the method depicted in Figure~\ref{fig:algo_fig}, but we get rid of the two phases, instead performing all steps simultaneously. The agent starts by randomly initializing the Laplacian representation, a set of options, and the main DDQN learner. As such, the options will initially maximize an essentially random intrinsic reward signal. As the Laplacian becomes more accurate the corresponding options are also adjusted. Importantly, such an approach also addresses another common criticism of Laplacian-based methods: that they are policy-dependent. It addresses this concern by being online since it simply tracks the changing policy learned by the agent. A more precise description of our algorithm is depicted in Algorithm \ref{algo:dceo_online}.

\paragraph{Empirical Analysis.} We evaluate our online algorithm in the same pixel-based environments we have been using so far.
As the results in Figure \ref{fig:fa_return_maximization_online} show, the online version of DCEO continues to be competitive or outperform the other baselines. Moreover, this fully online algorithm is even competitive with the two-phased algorithm, as shown in Figure \ref{fig:comparison_online_vs_stages} in Appendix~\ref{app:compare_dceo}. 
Because online DCEO performs similarly to the two-phased DCEO, and because it is simpler, we use the online DCEO algorithm in the rest of the paper, labeling its curves simply as DCEO.

\subsection{Exploration in the Face of Non-stationarity}
\label{sec:transfer}
An important benefit of using options for exploration is that, by encoding temporally extended behaviours into a set of options, the agent can later leverage a collection of diverse and purposeful behaviours in other tasks. This is particularly important in the face of non-stationary, or continual learning ~\citep{Khetarpal2020TowardsCR}, and is in direct contrast to several other exploration techniques. Methods such as count-based or error prediction-based methods are more tied to the agent's state visitation distribution and are not that flexible in the face of non-stationarity. It is often argued that options could allow agents to adapt more efficiently to non-stationarity through task decomposition, where a set of tasks share a similar structure. Here we show that options can also encode reusable exploratory behaviour and be leveraged for continual exploration.

To evaluate the discussed approaches in this setting, we introduce a variation of our \textit{Nine rooms} where the environment changes after a certain number of timesteps. We first consider the setting where only the agent's starting location and the goal location change. Importantly, the agent gets no information about this change as the goal is not visible in the input feature space and the agent does not get to reset any of its components. It simply must adapt online to the change and discover the new task. We also consider the much more challenging setting where the topology of the environment also changes---after the switch most of the hallways are closed, changing significantly the underlying graph of the environment, which is shown in Figure \ref{fig:non_stationary_envs} in Appendix \ref{app:non_stationary}. Because the options learned by DCEO are defined through the eigenfunctions of the graph Laplacian, which depend mainly on the environment's topology, this could arguably be a particularly difficult task for DCEO.

\begin{figure}[t]
     \centering
     \begin{subfigure}[b]{0.48\columnwidth}
         \centering
         \includegraphics[width=\columnwidth]{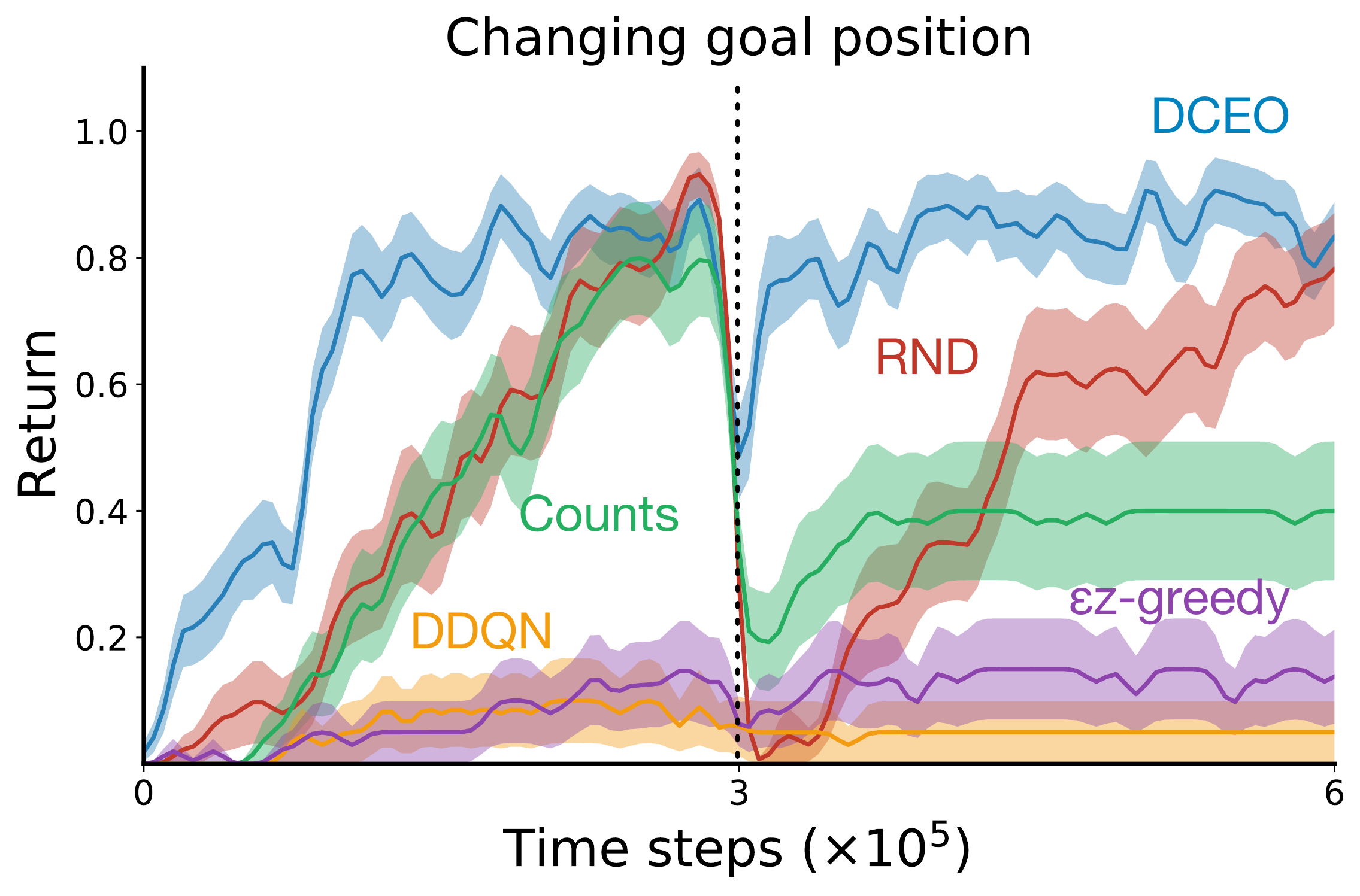}
     \end{subfigure}
     ~
     \begin{subfigure}[b]{0.49\columnwidth}
         \centering
         \includegraphics[width=\columnwidth]{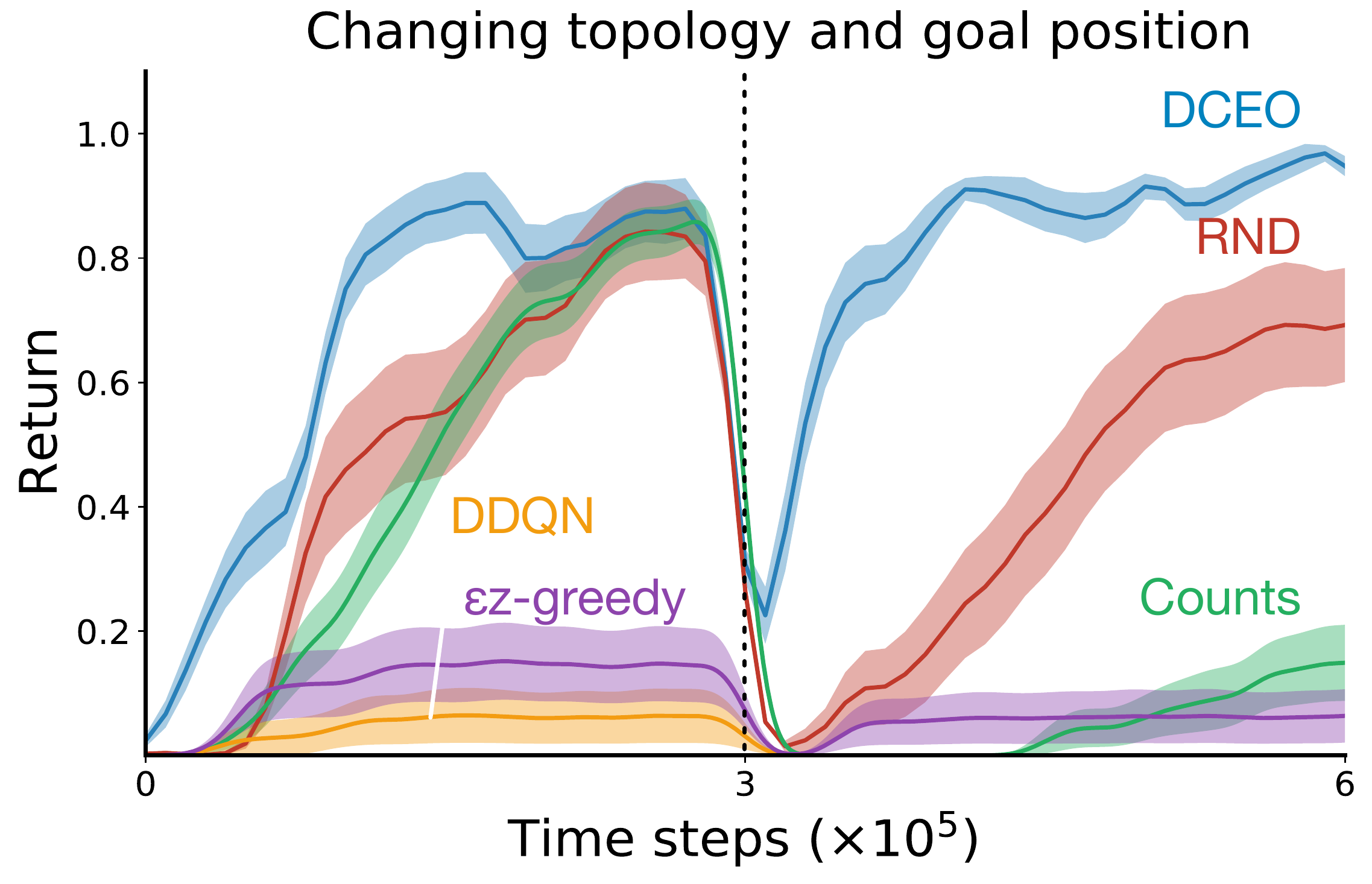}
     \end{subfigure}
        \caption{\textbf{Continual exploration} in  Nine rooms. Results show mean and standard deviation averaged over 30 random seeds.}
        \label{fig:non_stationarity}
\end{figure}

The empirical results are presented in Figure \ref{fig:non_stationarity}. In both problems we notice that DCEO adapts much more quickly compared to other baselines as the gap in performance is significantly increased after transfer. The results on the right panel are particularly exciting as they show that DCEO's performance remains robust to drastic changes in the environment's topology, also attesting to the benefits of an online method and of having reusable artifacts, since some options are still useful after the environment changes.

\subsection{Objects, Obstacles and Complex Interactions}

\begin{figure*}[h!]
     \centering
     \begin{subfigure}[b]{0.3\textwidth}
         \centering
         \includegraphics[width=\textwidth]{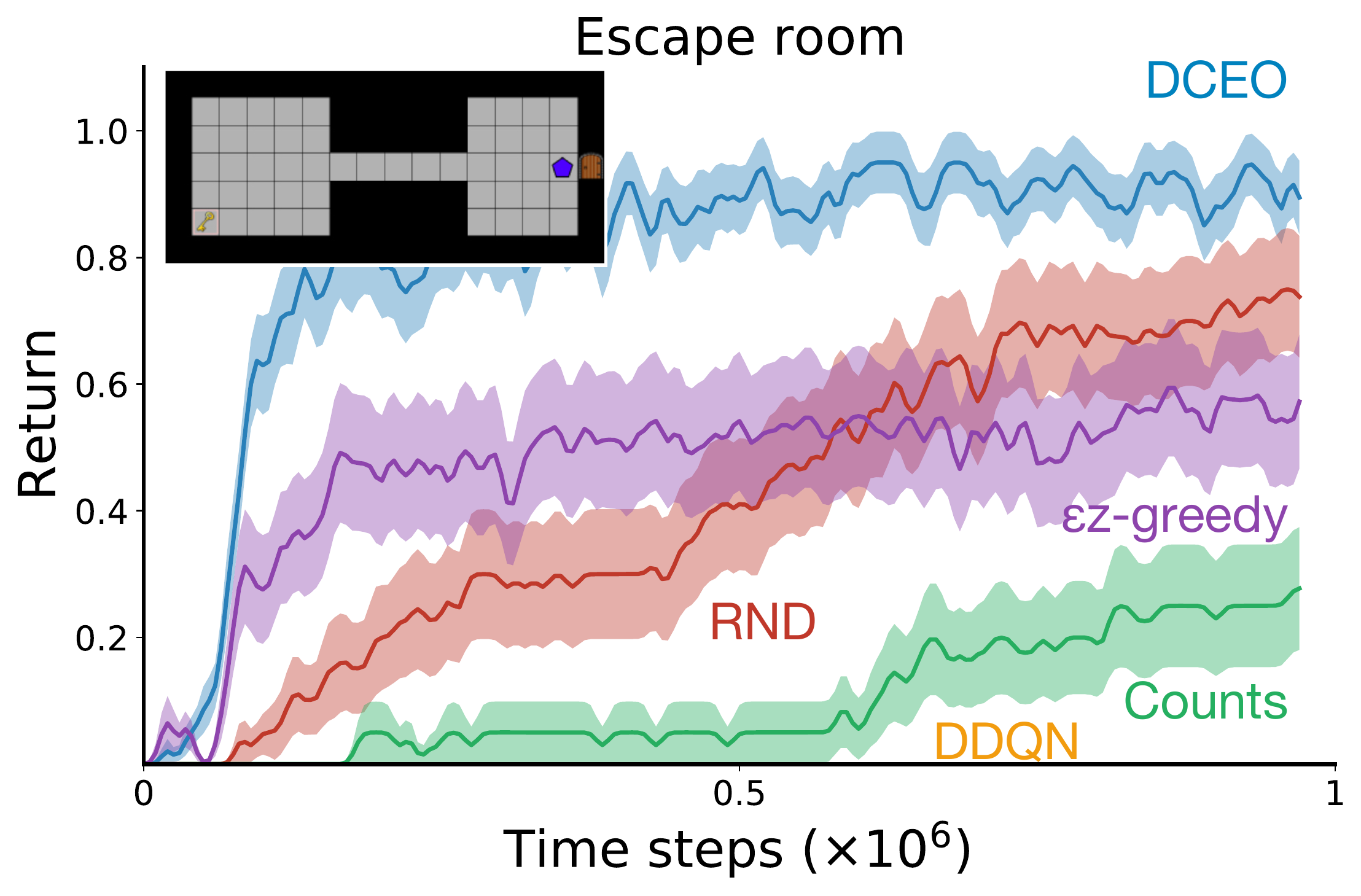}
     \end{subfigure}
     \hfill
     \begin{subfigure}[b]{0.3\textwidth}
         \centering
         \includegraphics[width=\textwidth]{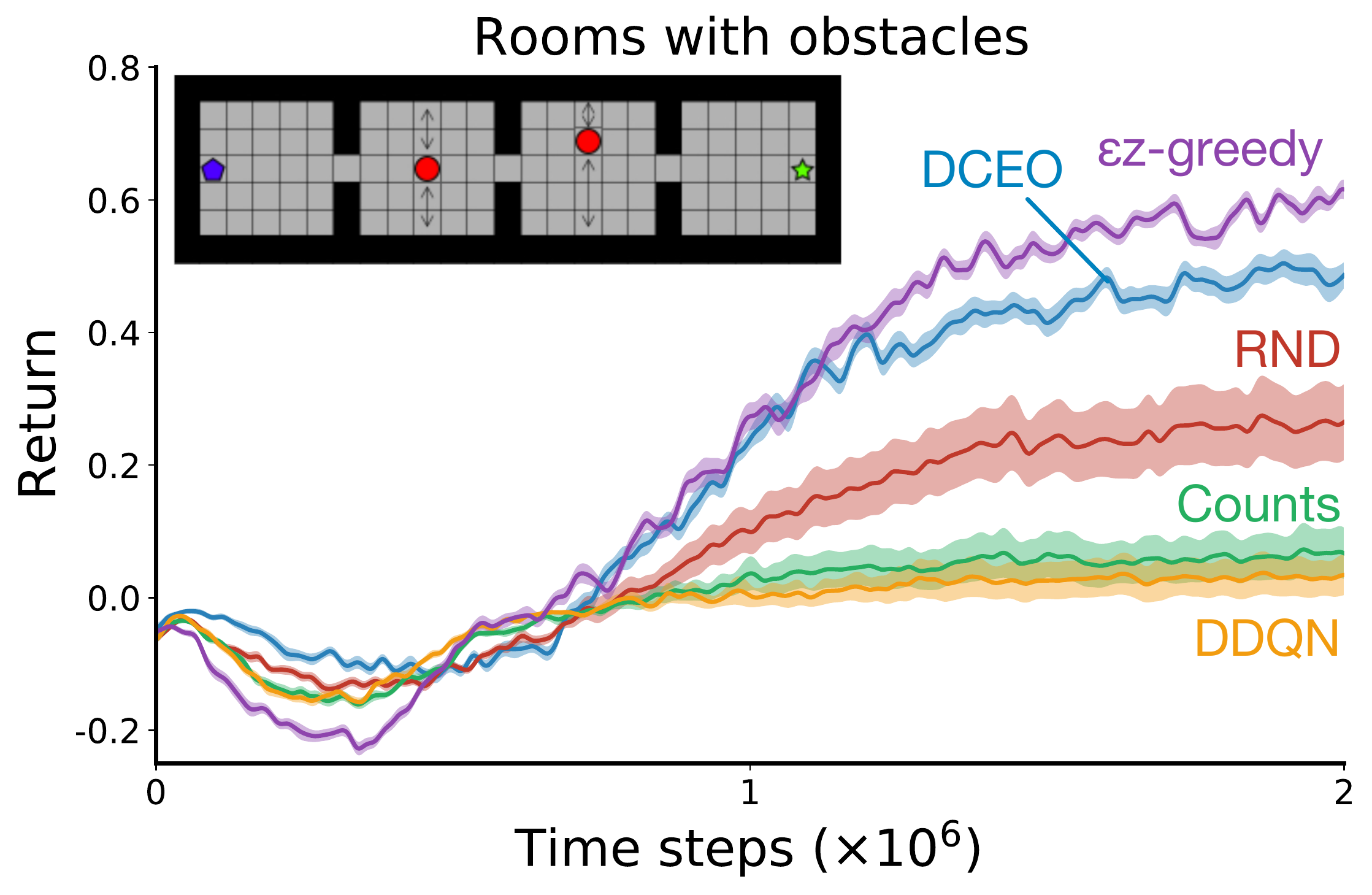}
     \end{subfigure}
     \hfill
     \begin{subfigure}[b]{0.3\textwidth}
         \centering
         \includegraphics[width=\textwidth]{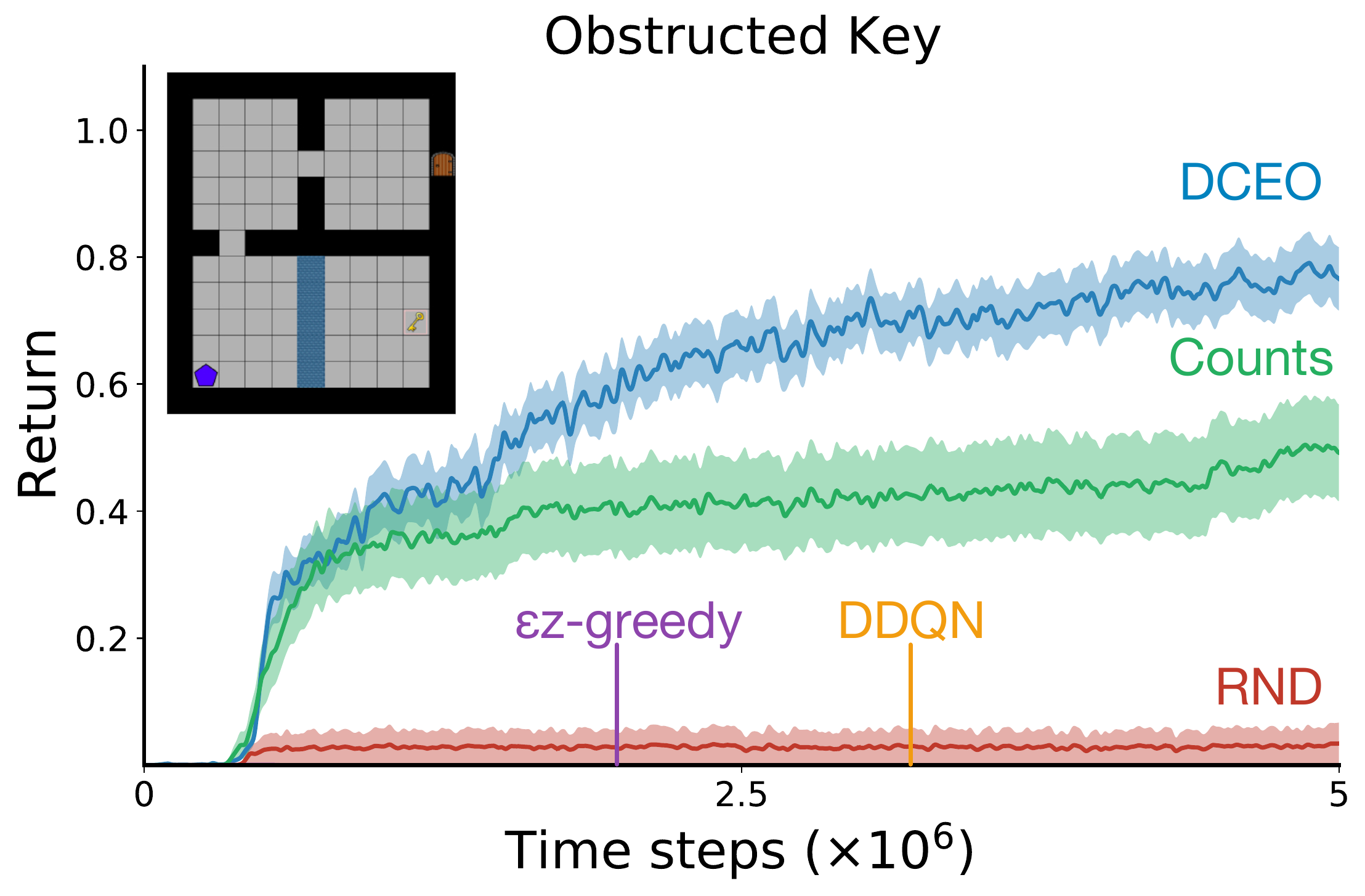}
     \end{subfigure}
        \caption{\textbf{Return maximization} in hard exploration domains that require different levels of abstraction to succeed. DCEO scales naturally to this challenging setting. Results show mean and standard deviation across 30 seeds. }
        \label{fig:fancy_fa_return_maximization}
\end{figure*}

So far we have investigated environments with different topologies and we have shown that our fully online DCEO algorithm is an effective method for exploration and that it has appealing properties in terms of continual exploration. We now apply the same algorithm to a set of challenging environments in which a combinatorial relation exists between the agent, objects, and their interactions. These environments are built on the Minigrid framework \citep{minigrid} and require different levels of abstraction to succesfully solve them. Importantly, all transitions are stochastic: the agent's action is overwritten by a random action with probability $0.15$. These environments are depicted as inset visualizations in Figure~\ref{fig:fancy_fa_return_maximization}.  

Specifically, in \textit{Escape room} the agent has to pick a key that is located in a separate room before returning to its starting location and escaping from the door. The agent receives a $+1$ only for escaping the room and $0$ otherwise. In \textit{Rooms with obstacles} the agent has to reach the goal location on the other end of the environment. In between are a set of dynamic obstacles with random spawning location that move up and down. If the agent collides with one of these obstacles it receives a $-1$ reward and the episode terminates. The agent receives a $+1$ only for reaching the goal. Finally, in \textit{Obstructed Key} the agent has to pick a key before escaping through a door located a few rooms away. However, the key is not directly accessible: the agent first has to break one (or more) tiles of the blue wall. The agent only receives a $+1$ for escaping the room through the door and $0$ otherwise. It is important to note that the last domain requires the agent to effectively change the environment topology. \looseness=-1

The results in Figure \ref{fig:fancy_fa_return_maximization} continue to attest to DCEO's generality as it also naturally extends to these challenging domains. In the most complex environment, \textit{Obstructed Key}, the gap in performance between DCEO and other baselines is the most significant. It is important to recall that the count-based baseline is still unfair as we give the agent access to perfect counts, not pseudo-counts, which maybe explains it outperforming the other baselines in this task.

Additionally, we use these experiments 
to provide some intuition as to what is encoded in the Laplacian representation in the presence of interactive objects. We present a visualization of the first (approximated) eigenfunction in Figure~\ref{fig:eigen_visualization} in Appendix \ref{app:visual_escape}. 
We witness an interesting property: the first eigenfunction seeks to pick the key and traverse, in the opposite way, the state space.
In this environment, when the agent picks up the key the observations change in a consistent way, leading to a whole new part in feature space. This important transition is naturally encoded in the Laplacian representation and it is essential for effectively exploring the environment. Furthermore, in Appendix \ref{app:additional_visuals} we present visualizations of how the Laplacian representation evolves as the agent covers the state space and learns to maximize reward. \looseness=-1

\section{Scaling Up Further}
\label{scale_up}
We now investigate the scalability of the Laplacian representation to even higher dimensional environments and to partial observability. In particular, we perform experiments on the Atari 2600 game Montezuma's Revenge through the Arcade Learning Environment \citep{bellemare2013arcade,machado2018revisiting}, a well-known hard exploration problem; and in the MiniWorld domain \citep{gym_miniworld}, a 3D navigation task with first-person view. In the latter we explore two tasks with different degrees of difficulty: the MiniWorld-FourRooms-v0, which recreates the Four rooms domain \citep{sutton1999between} in a 3D nagivation setting under partial-observability, and the more challenging MiniWorld-MyWayHomeSparse-v0, which recreates the classic VizDoom \citep{vizdoom} navigation task.

\begin{figure}[t]
     \centering
     \begin{subfigure}[b]{1.\columnwidth}
         \centering
         \includegraphics[width=\columnwidth]{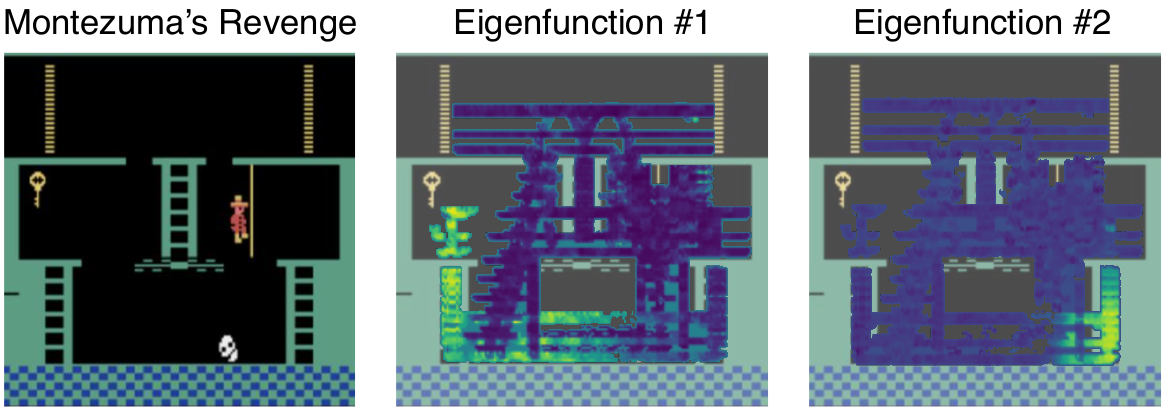}
     \end{subfigure}
     ~
     \begin{subfigure}[b]{1.\columnwidth}
         \centering
         \includegraphics[width=\columnwidth]{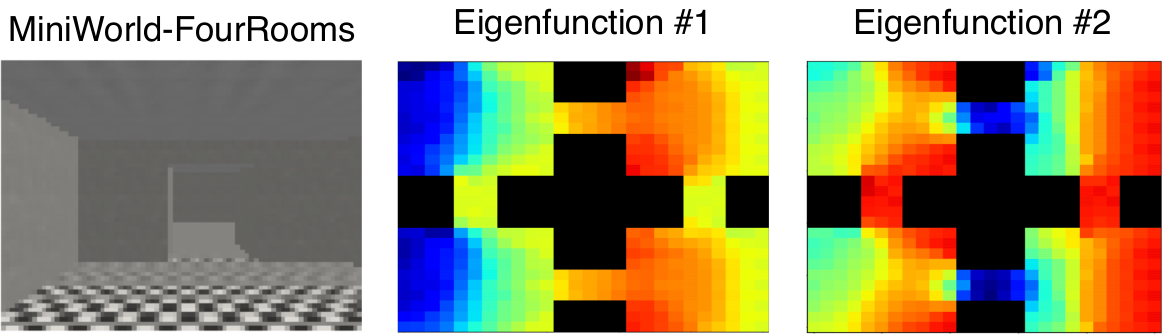}
     \end{subfigure}
        \caption{First two  eigenfunctions obtained by the generalized Laplacian. In Montezuma's Revenge, we plot the value of an eigenfunction for each of the agent's positions in the first room. In MW-FR, we plot the first two eigenfunctions for each of the agent's position in the map. The agent's point of view is a 3D first-person observation but we show values from a bird's eye view.}
        \label{fig:scale_up}
\end{figure}

We provide qualitative results in Fig.~\ref{fig:scale_up}, where we plot the values of the first two (approximate) eigenfunctions of the Laplacian representation for different observations. In these experiments, trajectories are obatined from random walks in the environemnt. For Montezuma's Revenge, this is achieved by coloring the pixels occupied by Panama Joe (the agent) in a set of randomly collected transitions. In MiniWorld-FourRooms (MW-FR), we spawn the agent at different locations and we save the value of the eigenfunction given the observation available at that position. The agent's orientation is fixed which allows for a bird's eye view plot of the eigenfunctions. \looseness=-1

Note that, in Montezuma's Revenge, the first eigenfunctions point to important stepping stones to get the key. Moreover, because meaningful locations are the first to be discovered, these results are quite different from previous results that required one to inspect dozens of eigenfunctions to find the interesting ones~\citep{machado2018eigenoption}. In MW-FR, as expected, the first eigenfunction seeks to traverse the observation space, leading the agent to cross the different rooms.\looseness=-1

These qualitative results continue to highlight the potential of Laplacian-based methods. They suggest that the first options to be discovered in such environments, without any domain knowledge, just by capturing the topology of the environment as experienced by the agent, would be meaningful. \looseness=-1

\begin{figure*}[h!]
     \centering
     \begin{subfigure}[b]{0.3\textwidth}
         \centering
         \includegraphics[width=\textwidth]{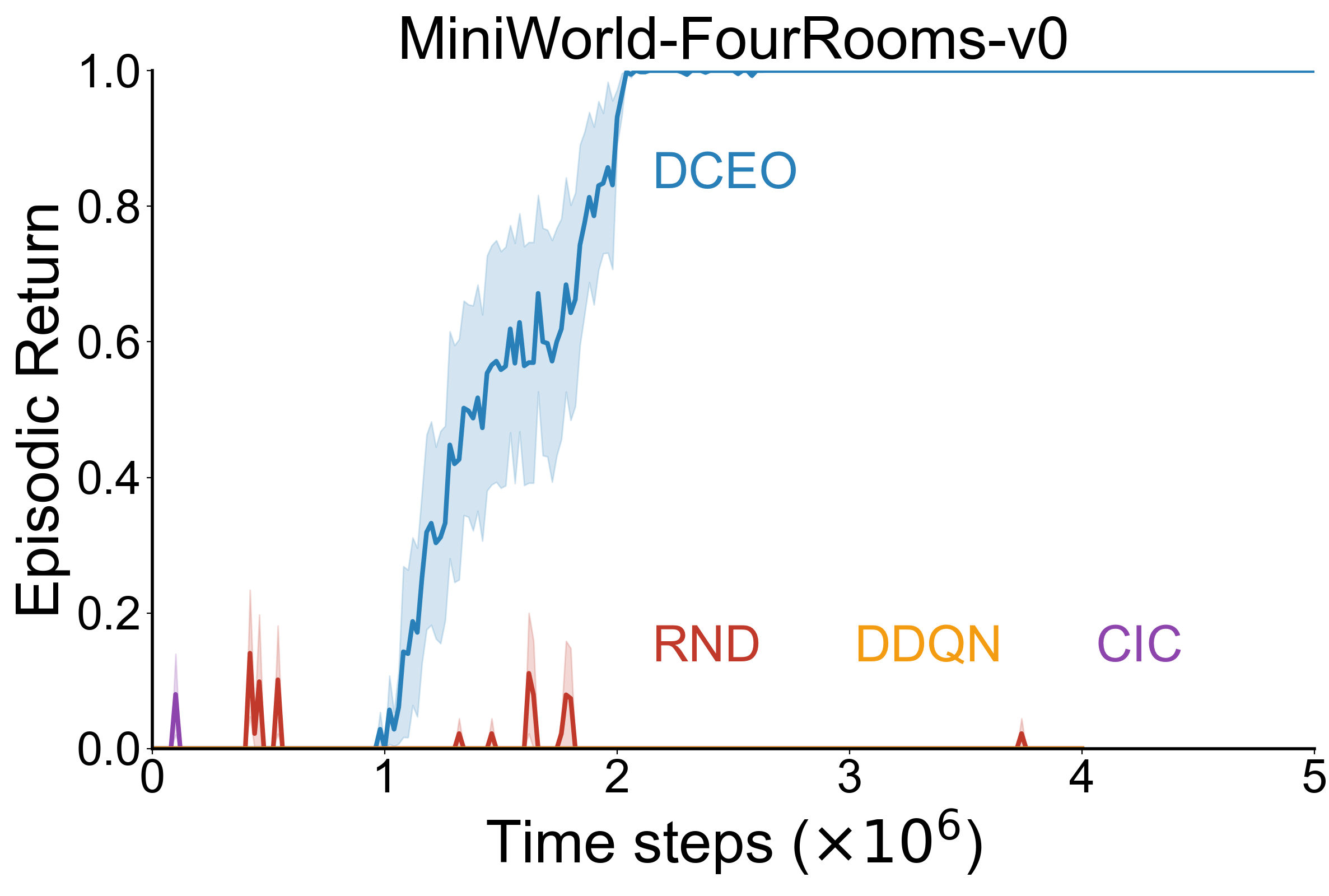}
     \end{subfigure}
     \hfill
     \begin{subfigure}[b]{0.3\textwidth}
         \centering
         \includegraphics[width=\textwidth]{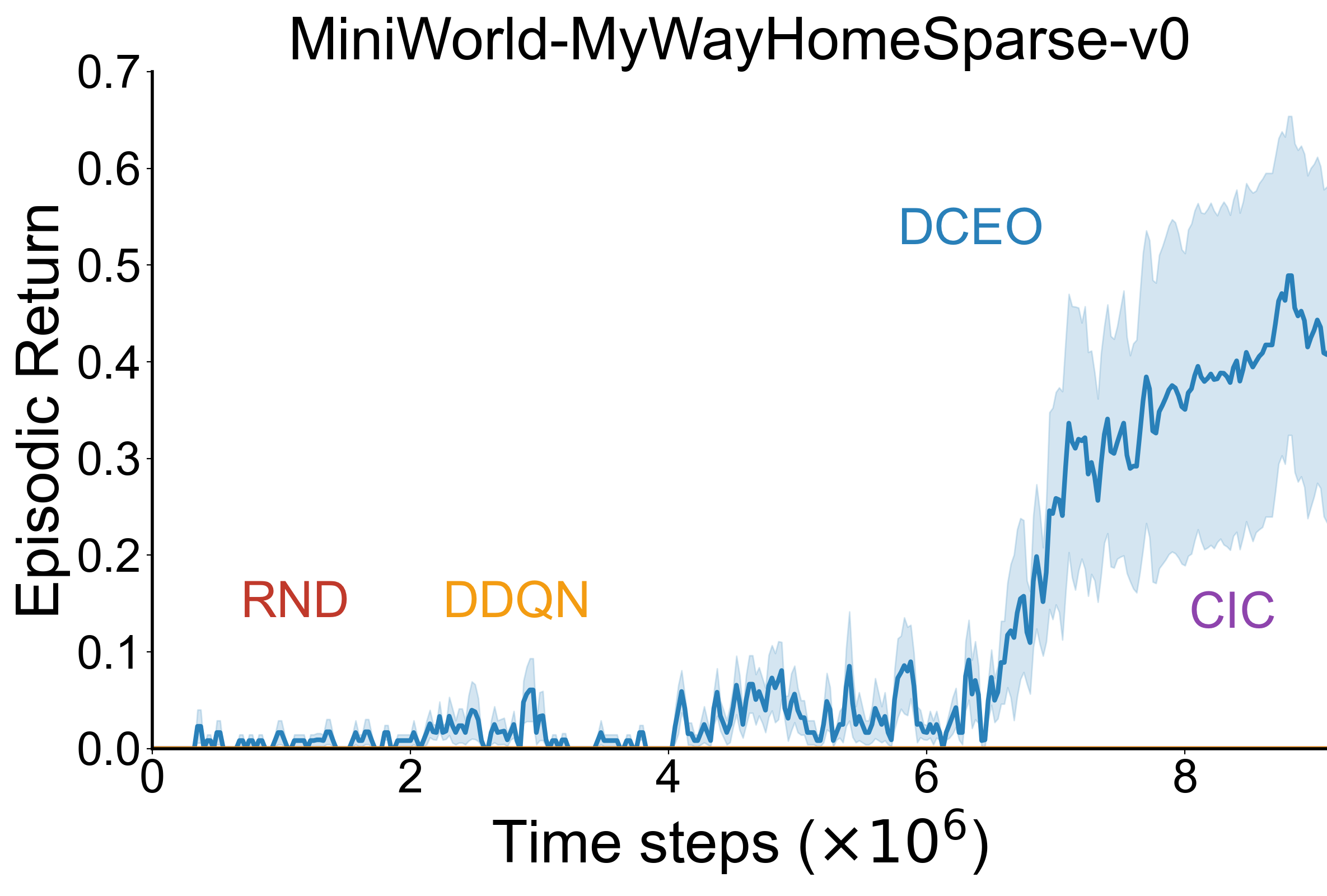}
     \end{subfigure}
     \hfill
     \begin{subfigure}[b]{0.3\textwidth}
         \centering
         \includegraphics[width=\textwidth]{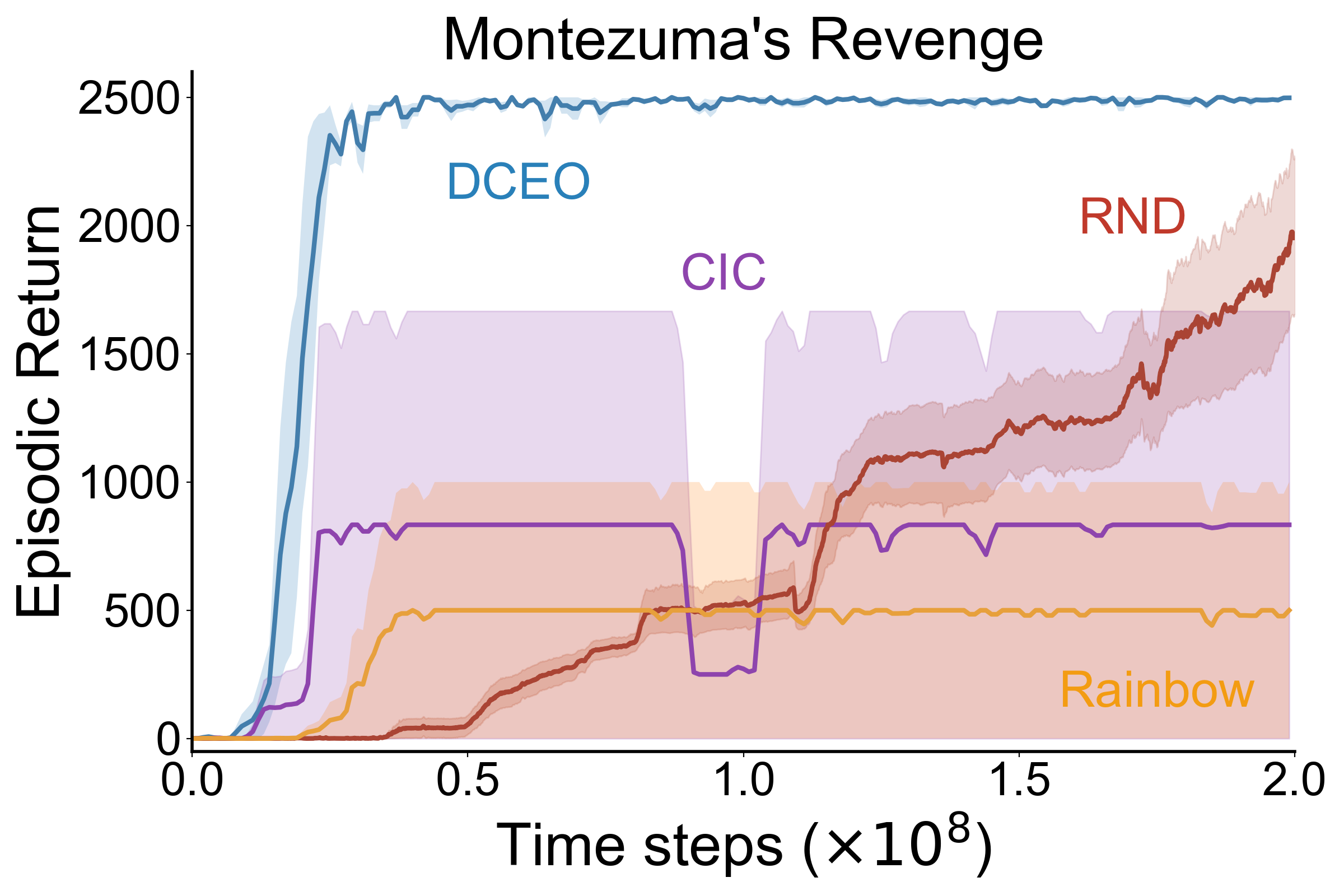}
     \end{subfigure}
        \caption{\textbf{Return maximization} in high dimensional hard exploration domains. The first two figures are tasks from the 3D Navigation domain MiniWorld, whereas the rightmost figure depict performance in the Atari 2600 game Montezuma's Revenge. DCEO once again scales naturally to these problems. Results show (from left to right) the mean and standard deviation across 10, 10 and 5 seeds.}
        \label{fig:large_scale}
\end{figure*}

We now verify, in a quantitative way, whether these discovered eigenfunctions can improve the exploration capabilities of a learning agent. Note once again that in DCEO the agent learns such eigenfunctions online through its own experience. We present these results in Figure \ref{fig:large_scale} where we compare DCEO to both RND and to a state-of-the-art diversity-based HRL algorithm, Contrastive Intrinsic Control (CIC) \citep{CIC}, which was shown to outperform a set of diversity-based skill learning algorithms in continuous control problems. We also include the learning algorithm used by DCEO but with a standard random exploration strategy---DDQN for the 3D Navigation tasks and Rainbow~\citep{hessel2018rainbow} for Montezuma's Revenge.

DCEO's benefits are very clear in the 3D navigation tasks, where it is the only algorithm to learn a policy able to consistently accumulate positive rewards. In Montezuma's Revenge, DCEO reaches a score of $2500$ in only a few million steps. Note that, when introduced, RND was run for 2 billion timesteps, and within that timeframe it surpasses DCEO's performance in 200 million steps; we did not have the resources to evaluate DCEO in 2 billion steps.

\section{Conclusion}
In this paper we have introduced a scalable and generally applicable algorithm for Laplacian-based option discovery.  Through a series of improvements, we have extended a tabular approach into an online method fully compatible with deep function approximation. We have proposed an effective strategy for incorporating option discovery and reward maximization and we have shown that our algorithm performs significantly better when compared to several state-of-of-the-art baselines on a wide variety of environments and settings. This is the first time that a Laplacian-based method has done so. The results in non-stationarity environments are particularly promising as they highlight the benefits of options and they suggest a new path for continual exploration. \looseness=-1

There remains a variety of research directions for future improvement. The integration in our approach is minimal by design. We do not learn option-value functions for credit assignment, we do not leverage the auxiliary task effect for representation learning nor we use options for planning. All of these directions capture long-held promises of the options framework: we believe that the same overarching Laplacian-based algorithm could be a way to achieve them.

\section*{Acknowledgements}
The authors would like to thank Adam White, Andre Barreto, Tom Zahavy, Michael Bowling, Diana Borsa and Doina Precup for constructive feedback throughout this project, Andy Patterson, Josh Davidson and the whole DeepMind Alberta team for helpful and inspiring discussions. A special thanks to Kaixin Wang for sharing the code of the generalized Laplacian and Kris de Asis for the availability of the Rubik's Cube 2x2 implementation. The authors would also like to thank NSERC for partially funding this research.

% \clearpage

\bibliography{main}

\begin{thebibliography}{62}
\providecommand{\natexlab}[1]{#1}
\providecommand{\url}[1]{\texttt{#1}}
\expandafter\ifx\csname urlstyle\endcsname\relax
  \providecommand{\doi}[1]{doi: #1}\else
  \providecommand{\doi}{doi: \begingroup \urlstyle{rm}\Url}\fi

\bibitem[Agostinelli et~al.(2019)Agostinelli, McAleer, Shmakov, and
  Baldi]{agostinelli2019solving}
Agostinelli, F., McAleer, S., Shmakov, A., and Baldi, P.
\newblock Solving the {R}ubik’s cube with deep reinforcement learning and
  search.
\newblock \emph{Nature Machine Intelligence}, 1:\penalty0 356--363, 2019.

\bibitem[Bacon et~al.(2017)Bacon, Harb, and Precup]{bacon2017option}
Bacon, P.-L., Harb, J., and Precup, D.
\newblock The option-critic architecture.
\newblock In \emph{AAAI Conference on Artificial Intelligence}, 2017.

\bibitem[Bagaria \& Konidaris(2020)Bagaria and Konidaris]{bagaria2020option}
Bagaria, A. and Konidaris, G.
\newblock Option discovery using deep skill chaining.
\newblock In \emph{International Conference on Learning Representations}, 2020.

\bibitem[Bar et~al.(2020)Bar, Talmon, and Meir]{bar2020option}
Bar, A., Talmon, R., and Meir, R.
\newblock Option discovery in the absence of rewards with manifold analysis.
\newblock In \emph{International Conference on Machine Learning}, 2020.

\bibitem[Bellemare et~al.(2013)Bellemare, Naddaf, Veness, and
  Bowling]{bellemare2013arcade}
Bellemare, M.~G., Naddaf, Y., Veness, J., and Bowling, M.
\newblock The {A}rcade {L}earning {E}nvironment: {A}n evaluation platform for
  general agents.
\newblock \emph{Journal of Artificial Intelligence Research}, 47:\penalty0
  253--279, 2013.

\bibitem[Bellemare et~al.(2016)Bellemare, Srinivasan, Ostrovski, Schaul,
  Saxton, and Munos]{bellemare2016unifying}
Bellemare, M.~G., Srinivasan, S., Ostrovski, G., Schaul, T., Saxton, D., and
  Munos, R.
\newblock Unifying count-based exploration and intrinsic motivation.
\newblock In \emph{{Neural Information Processing Systems}}, 2016.

\bibitem[Bellemare et~al.(2020)Bellemare, Candido, Castro, Gong, Machado,
  Moitra, Ponda, and Wang]{bellemare2020autonomous}
Bellemare, M.~G., Candido, S., Castro, P.~S., Gong, J., Machado, M.~C., Moitra,
  S., Ponda, S.~S., and Wang, Z.
\newblock Autonomous navigation of stratospheric balloons using reinforcement
  learning.
\newblock \emph{Nature}, 588:\penalty0 77--82, 2020.

\bibitem[Burda et~al.(2019)Burda, Edwards, Storkey, and
  Klimov]{burda2019exploration}
Burda, Y., Edwards, H., Storkey, A., and Klimov, O.
\newblock Exploration by random network distillation.
\newblock In \emph{International Conference on Learning Representations}, 2019.

\bibitem[Chevalier-Boisvert(2018)]{gym_miniworld}
Chevalier-Boisvert, M.
\newblock Miniworld: Minimalistic 3d environment for {RL} \& robotics research.
\newblock \url{https://github.com/maximecb/gym-miniworld}, 2018.

\bibitem[Chevalier-Boisvert et~al.(2018)Chevalier-Boisvert, Willems, and
  Pal]{minigrid}
Chevalier-Boisvert, M., Willems, L., and Pal, S.
\newblock Minimalistic gridworld environment for {OpenAI Gym}.
\newblock \url{https://github.com/Farama-Foundation/Minigrid}, 2018.

\bibitem[Chung(1997)]{chung1997spectral}
Chung, F. R.~K.
\newblock \emph{Spectral Graph Theory}.
\newblock Conference Board of the Mathematical Sciences, 1997.

\bibitem[Dabney et~al.(2021)Dabney, Ostrovski, and
  Barreto]{dabney2021temporally}
Dabney, W., Ostrovski, G., and Barreto, A.
\newblock Temporally-extended {\(\epsilon\)}-greedy exploration.
\newblock In \emph{International Conference on Learning Representations}, 2021.

\bibitem[Dayan \& Hinton(1992)Dayan and Hinton]{dayan1992feudal}
Dayan, P. and Hinton, G.~E.
\newblock Feudal reinforcement learning.
\newblock In \emph{Neural Information Processing Systems}, 1992.

\bibitem[de~Asis(2018)]{kris_cube}
de~Asis, K.
\newblock py222.
\newblock \url{https://github.com/MeepMoop/py222}, 2018.

\bibitem[Deng et~al.(2022)Deng, Shi, Zhang, Cui, Lu, and Zhu]{deng20222neural}
Deng, Z., Shi, J., Zhang, H., Cui, P., Lu, C., and Zhu, J.
\newblock Neural eigenfunctions are structured representation learners.
\newblock \emph{CoRR}, abs/2210.12637, 2022.

\bibitem[Ecoffet et~al.(2021)Ecoffet, Huizinga, Lehman, Stanley, and
  Clune]{ecoffet2021first}
Ecoffet, A., Huizinga, J., Lehman, J., Stanley, K.~O., and Clune, J.
\newblock First return, then explore.
\newblock \emph{Nature}, 590:\penalty0 580--586, 2021.

\bibitem[Erraqabi et~al.(2022)Erraqabi, Machado, Zhao, Sukhbaatar, Lazaric,
  Denoyer, and Bengio]{erraqabi2021temporal}
Erraqabi, A., Machado, M.~C., Zhao, M., Sukhbaatar, S., Lazaric, A., Denoyer,
  L., and Bengio, Y.
\newblock Temporal abstractions-augmented temporally contrastive learning: An
  alternative to the {L}aplacian in {RL}.
\newblock In \emph{Conference on Uncertainty in Artificial Intelligence}, 2022.

\bibitem[Eysenbach et~al.(2019)Eysenbach, Gupta, Ibarz, and
  Levine]{eysenbach2019diversity}
Eysenbach, B., Gupta, A., Ibarz, J., and Levine, S.
\newblock Diversity is all you need: {L}earning skills without a reward
  function.
\newblock In \emph{International Conference on Learning Representations}, 2019.

\bibitem[Farebrother et~al.(2023)Farebrother, Greaves, Agarwal, Lan, Goroshin,
  Castro, and Bellemare]{farebrother2023protovalue}
Farebrother, J., Greaves, J., Agarwal, R., Lan, C.~L., Goroshin, R., Castro,
  P.~S., and Bellemare, M.~G.
\newblock Proto-value networks: Scaling representation learning with auxiliary
  tasks.
\newblock In \emph{International Conference on Learning Representations}, 2023.

\bibitem[Gregor et~al.(2017)Gregor, Rezende, and Wierstra]{gregor16variational}
Gregor, K., Rezende, D., and Wierstra, D.
\newblock Variational intrinsic control.
\newblock In \emph{International Conference on Learning Representations,
  Workshop track}, 2017.

\bibitem[Harb et~al.(2018)Harb, Bacon, Klissarov, and Precup]{harb2018when}
Harb, J., Bacon, P., Klissarov, M., and Precup, D.
\newblock When waiting is not an option: Learning options with a deliberation
  cost.
\newblock In \emph{{AAAI} Conference on Artificial Intelligence}, 2018.

\bibitem[Hessel et~al.(2018)Hessel, Modayil, van Hasselt, Schaul, Ostrovski,
  Dabney, Horgan, Piot, Azar, and Silver]{hessel2018rainbow}
Hessel, M., Modayil, J., van Hasselt, H., Schaul, T., Ostrovski, G., Dabney,
  W., Horgan, D., Piot, B., Azar, M.~G., and Silver, D.
\newblock Rainbow: Combining improvements in deep reinforcement learning.
\newblock In \emph{{AAAI} Conference on Artificial Intelligence}, 2018.

\bibitem[Jinnai et~al.(2019)Jinnai, Park, Abel, and
  Konidaris]{jinnai2019discovering}
Jinnai, Y., Park, J.~W., Abel, D., and Konidaris, G.
\newblock Discovering options for exploration by minimizing cover time.
\newblock In \emph{International Conference on Machine Learning}, 2019.

\bibitem[Jinnai et~al.(2020)Jinnai, Park, Machado, and
  Konidaris]{jinnai2020exploration}
Jinnai, Y., Park, J.~W., Machado, M.~C., and Konidaris, G.
\newblock Exploration in reinforcement learning with deep covering options.
\newblock In \emph{International Conference on Learning Representations}, 2020.

\bibitem[Kempka et~al.(2016)Kempka, Wydmuch, Runc, Toczek, and
  Jaśkowski]{vizdoom}
Kempka, M., Wydmuch, M., Runc, G., Toczek, J., and Jaśkowski, W.
\newblock {ViZDoom}: A {D}oom-based {AI} research platform for visual
  reinforcement learning.
\newblock In \emph{IEEE Conference on Computational Intelligence and Games},
  2016.

\bibitem[Khetarpal et~al.(2020)Khetarpal, Klissarov, Chevalier{-}Boisvert,
  Bacon, and Precup]{ioc}
Khetarpal, K., Klissarov, M., Chevalier{-}Boisvert, M., Bacon, P., and Precup,
  D.
\newblock Options of interest: Temporal abstraction with interest functions.
\newblock In \emph{{AAAI} Conference on Artificial Intelligence}, 2020.

\bibitem[Khetarpal et~al.(2022)Khetarpal, Riemer, Rish, and
  Precup]{Khetarpal2020TowardsCR}
Khetarpal, K., Riemer, M., Rish, I., and Precup, D.
\newblock Towards continual reinforcement learning: A review and perspectives.
\newblock \emph{Journal of Artificial Intelligence Research}, 75:\penalty0
  1401--1476, 2022.

\bibitem[Kim et~al.(2021)Kim, Park, and Kim]{kim2021unsupervised}
Kim, J., Park, S., and Kim, G.
\newblock Unsupervised skill discovery with bottleneck option learning.
\newblock In \emph{International Conference on Machine Learning}, 2021.

\bibitem[Klissarov \& Precup(2020)Klissarov and Precup]{klissarov2020reward}
Klissarov, M. and Precup, D.
\newblock Reward propagation using graph convolutional networks.
\newblock In \emph{Neural Information Processing Systems}, 2020.

\bibitem[Klissarov \& Precup(2021)Klissarov and Precup]{NEURIPS2021_24cceab7}
Klissarov, M. and Precup, D.
\newblock Flexible option learning.
\newblock In \emph{Neural Information Processing Systems}, 2021.

\bibitem[Konidaris \& Barto(2009)Konidaris and Barto]{konidaris2009skill}
Konidaris, G. and Barto, A.
\newblock Skill discovery in continuous reinforcement learning domains using
  skill chaining.
\newblock In \emph{Neural Information Processing Systems}, 2009.

\bibitem[Koren(2005)]{koren2005drawing}
Koren, Y.
\newblock Drawing graphs by eigenvectors: Theory and practice.
\newblock \emph{Computers \& Mathematics with Applications}, 49\penalty0
  (11):\penalty0 1867--1888, 2005.

\bibitem[Laskin et~al.(2022)Laskin, Liu, Peng, Yarats, Rajeswaran, and
  Abbeel]{CIC}
Laskin, M., Liu, H., Peng, X.~B., Yarats, D., Rajeswaran, A., and Abbeel, P.
\newblock {CIC}: Contrastive intrinsic control for unsupervised skill
  discovery.
\newblock In \emph{Neural Information Processing Systems}, 2022.

\bibitem[Lin(1992)]{lin1992self}
Lin, L.-J.
\newblock Self-improving reactive agents based on reinforcement learning,
  planning and teaching.
\newblock \emph{Machine Learning}, 8\penalty0 (3–4):\penalty0 293–321,
  1992.

\bibitem[Lyle et~al.(2021)Lyle, Rowland, Ostrovski, and Dabney]{lyle2021effect}
Lyle, C., Rowland, M., Ostrovski, G., and Dabney, W.
\newblock On the effect of auxiliary tasks on representation dynamics.
\newblock In \emph{International Conference on Artificial Intelligence and
  Statistics}, 2021.

\bibitem[Machado \& Bowling(2016)Machado and Bowling]{machado16learning}
Machado, M.~C. and Bowling, M.
\newblock Learning purposeful behaviour in the absence of rewards.
\newblock In \emph{ICML Workshop on Abstraction in Reinforcement Learning},
  2016.

\bibitem[Machado et~al.(2017)Machado, Bellemare, and
  Bowling]{machado2017laplacian}
Machado, M.~C., Bellemare, M.~G., and Bowling, M.
\newblock A {L}aplacian framework for option discovery in reinforcement
  learning.
\newblock In \emph{International Conference on Machine Learning}, 2017.

\bibitem[Machado et~al.(2018{\natexlab{a}})Machado, Bellemare, Talvitie,
  Veness, Hausknecht, and Bowling]{machado2018revisiting}
Machado, M.~C., Bellemare, M.~G., Talvitie, E., Veness, J., Hausknecht, M., and
  Bowling, M.
\newblock Revisiting the {A}rcade {L}earning {E}nvironment: {E}valuation
  protocols and open problems for general agents.
\newblock \emph{{Journal of Artificial Intelligence Research}}, 61:\penalty0
  523--562, 2018{\natexlab{a}}.

\bibitem[Machado et~al.(2018{\natexlab{b}})Machado, Rosenbaum, Guo, Liu,
  Tesauro, and Campbell]{machado2018eigenoption}
Machado, M.~C., Rosenbaum, C., Guo, X., Liu, M., Tesauro, G., and Campbell, M.
\newblock Eigenoption discovery through the deep successor representation.
\newblock In \emph{International Conference on Learning Representations},
  2018{\natexlab{b}}.

\bibitem[Machado et~al.(2023)Machado, Barreto, Precup, and
  Bowling]{machado21temporal}
Machado, M.~C., Barreto, A., Precup, D., and Bowling, M.
\newblock Temporal abstraction in reinforcement learning with the successor
  representation.
\newblock \emph{Journal of Machine Learning Research}, 24:\penalty0 1--69,
  2023.

\bibitem[Mahadevan(2005)]{mahadevan2005proto}
Mahadevan, S.
\newblock Proto-value functions: Developmental reinforcement learning.
\newblock In \emph{International Conference on Machine Learning}, 2005.

\bibitem[Mahadevan \& Maggioni(2007)Mahadevan and Maggioni]{mahadevan2007proto}
Mahadevan, S. and Maggioni, M.
\newblock Proto-value functions: {A} {L}aplacian framework for learning
  representation and control in markov decision processes.
\newblock \emph{{Journal of Machine Learning Research}}, 8:\penalty0
  2169--2231, 2007.

\bibitem[Mall et~al.(2013)Mall, Langone, and Suykens]{mall}
Mall, R., Langone, R., and Suykens, J.
\newblock Kernel spectral clustering for big data networks.
\newblock \emph{Entropy}, 2013.

\bibitem[Mnih et~al.(2013)Mnih, Kavukcuoglu, Silver, Graves, Antonoglou,
  Wierstra, and Riedmiller]{mnih2013playing}
Mnih, V., Kavukcuoglu, K., Silver, D., Graves, A., Antonoglou, I., Wierstra,
  D., and Riedmiller, M.
\newblock Playing {A}tari with deep reinforcement learning.
\newblock In \emph{NeurIPS Deep Learning Workshop}. 2013.

\bibitem[Mnih et~al.(2015)Mnih, Kavukcuoglu, Silver, Rusu, Veness, Bellemare,
  Graves, Riedmiller, Fidjeland, Ostrovski, Petersen, Beattie, Sadik,
  Antonoglou, King, Kumaran, Wierstra, Legg, and Hassabis]{mnih2015human}
Mnih, V., Kavukcuoglu, K., Silver, D., Rusu, A.~A., Veness, J., Bellemare,
  M.~G., Graves, A., Riedmiller, M., Fidjeland, A.~K., Ostrovski, G., Petersen,
  S., Beattie, C., Sadik, A., Antonoglou, I., King, H., Kumaran, D., Wierstra,
  D., Legg, S., and Hassabis, D.
\newblock Human-level control through deep reinforcement learning.
\newblock \emph{Nature}, 518:\penalty0 529--533, 2015.

\bibitem[Nadler et~al.(2006)Nadler, Lafon, Coifman, and
  Kevrekidis]{nadler2006diffusion}
Nadler, B., Lafon, S., Coifman, R.~R., and Kevrekidis, I.~G.
\newblock Diffusion maps, spectral clustering and reaction coordinates of
  dynamical systems.
\newblock \emph{Applied and Computational Harmonic Analysis}, 21\penalty0
  (1):\penalty0 113--127, 2006.

\bibitem[Oja(1982)]{oja-simplified-neuron-model-1982}
Oja, E.
\newblock Simplified neuron model as a principal component analyzer.
\newblock \emph{Journal of Mathematical Biology}, 1982.

\bibitem[Park et~al.(2022)Park, Choi, Kim, Lee, and Kim]{park202lipschitz}
Park, S., Choi, J., Kim, J., Lee, H., and Kim, G.
\newblock Lipschitz-constrained unsupervised skill discovery.
\newblock In \emph{International Conference on Learning Representations}, 2022.

\bibitem[Pfau et~al.(2019)Pfau, Petersen, Agarwal, Barrett, and
  Stachenfeld]{pfau2018spectral}
Pfau, D., Petersen, S., Agarwal, A., Barrett, D. G.~T., and Stachenfeld, K.~L.
\newblock Spectral inference networks: Unifying deep and spectral learning.
\newblock In \emph{International Conference on Learning Representations}, 2019.

\bibitem[Precup(2000)]{precup00temporal}
Precup, D.
\newblock \emph{Temporal Abstraction in Reinforcement Learning}.
\newblock PhD thesis, University of Massachusetts Amherst, 2000.

\bibitem[Ren et~al.(2023)Ren, Zhang, Lee, Gonzalez, Schuurmans, and
  Dai]{ren2023spectral}
Ren, T., Zhang, T., Lee, L., Gonzalez, J.~E., Schuurmans, D., and Dai, B.
\newblock Spectral decomposition representation for reinforcement learning.
\newblock In \emph{International Conference on Learning Representations}, 2023.

\bibitem[Smith et~al.(2018)Smith, van Hoof, and Pineau]{smith2018inference}
Smith, M., van Hoof, H., and Pineau, J.
\newblock An inference-based policy gradient method for learning options.
\newblock In \emph{International Conference on Machine Learning}, 2018.

\bibitem[Sutton et~al.(1999)Sutton, Precup, and Singh]{sutton1999between}
Sutton, R., Precup, D., and Singh, S.
\newblock Between {MDP}s and semi-{MDP}s: A framework for temporal abstraction
  in reinforcement learning.
\newblock \emph{Artificial Intelligence}, 112\penalty0 (1–2):\penalty0 181 --
  211, 1999.

\bibitem[Sutton et~al.(2011)Sutton, Modayil, Delp, Degris, Pilarski, White, and
  Precup]{sutton2011horde}
Sutton, R., Modayil, J., Delp, M., Degris, T., Pilarski, P.~M., White, A., and
  Precup, D.
\newblock Horde: A scalable real-time architecture for learning knowledge from
  unsupervised sensorimotor interaction.
\newblock In \emph{International Conference on Autonomous Agents {\&}
  Multiagent Systems}, 2011.

\bibitem[Sutton \& Barto(2018)Sutton and Barto]{sutton2018reinforcement}
Sutton, R.~S. and Barto, A.~G.
\newblock \emph{Reinforcement Learning: An Introduction}.
\newblock The MIT Press, second edition, 2018.

\bibitem[Touati et~al.(2023)Touati, Rapin, and Ollivier]{touati2023does}
Touati, A., Rapin, J., and Ollivier, Y.
\newblock Does zero-shot reinforcement learning exist?
\newblock In \emph{International Conference on Learning Representations}, 2023.

\bibitem[van Hasselt et~al.(2016)van Hasselt, Guez, and
  Silver]{hasselt2016deep}
van Hasselt, H., Guez, A., and Silver, D.
\newblock Deep reinforcement learning with double {Q}-learning.
\newblock In \emph{{AAAI} Conference on Artificial Intelligence}, 2016.

\bibitem[Vezhnevets et~al.(2017)Vezhnevets, Osindero, Schaul, Heess, Jaderberg,
  Silver, and Kavukcuoglu]{vezhnevets2017feudal}
Vezhnevets, A., Osindero, S., Schaul, T., Heess, N., Jaderberg, M., Silver, D.,
  and Kavukcuoglu, K.
\newblock Feudal networks for hierarchical reinforcement learning.
\newblock In \emph{International Conference on Machine Learning}, 2017.

\bibitem[Wang et~al.(2021)Wang, Zhou, Zhang, Shao, Hooi, and
  Feng]{wang2021towards}
Wang, K., Zhou, K., Zhang, Q., Shao, J., Hooi, B., and Feng, J.
\newblock Towards better {L}aplacian representation in reinforcement learning
  with generalized graph drawing.
\newblock In \emph{International Conference on Machine Learning}, 2021.

\bibitem[Watkins \& Dayan(1992)Watkins and Dayan]{watkins1992q}
Watkins, C. J. C.~H. and Dayan, P.
\newblock Technical note: \cal {Q}-learning.
\newblock \emph{Machine Learning}, 8\penalty0 (3-4), May 1992.

\bibitem[Wu et~al.(2019)Wu, Tucker, and Nachum]{wu2019laplacian}
Wu, Y., Tucker, G., and Nachum, O.
\newblock The {L}aplacian in {RL:} learning representations with efficient
  approximations.
\newblock In \emph{International Conference on Learning Representations}, 2019.

\bibitem[Wulfmeier et~al.(2021)Wulfmeier, Rao, Hafner, Lampe, Abdolmaleki,
  Hertweck, Neunert, Tirumala, Siegel, Heess, and
  Riedmiller]{wulfmeier2021data}
Wulfmeier, M., Rao, D., Hafner, R., Lampe, T., Abdolmaleki, A., Hertweck, T.,
  Neunert, M., Tirumala, D., Siegel, N., Heess, N., and Riedmiller, M.
\newblock Data-efficient hindsight off-policy option learning.
\newblock In \emph{International Conference on Machine Learning}, 2021.

\end{thebibliography}
\bibliographystyle{icml2023} 

\clearpage
\onecolumn
\appendix

\section{Related Work}

The graph Laplacian has been introduced in RL through the Proto-Value Functions (PVFs) framework~\citep{mahadevan2005proto,mahadevan2007proto} in which the eigenfunctions of the Laplacian are used as a solution to the problem of representation learning. The same quantity was leveraged to learn eigenoptions \cite{machado16learning,machado2017laplacian}, that is, options that follow the eigenfunctions of the graph Laplacian. The full eigenspectrum was later used by \citet{bar2020option} to define options that leverage the diffusion distance \citep{nadler2006diffusion}. Another line of work defines options through a single eigenfunction of the Laplacian and show that options obtained by maximizing this vector lead to better state coverage \citep{jinnai2019discovering}. An important limitation of these approaches is that they are not naturally scalable because they are evaluated or derived using the true eigenfunctions of the graph Laplacian, which requires performing an eigendecomposition on an $|\mathscr{S}| \times |\mathscr{S}|$ matrix that is not readily available.

In machine learning, there are various approaches that advocate for avoiding the costly eigendecomposition operation and approximating the eigenfunctions. This line of work dates back to Oja’s rule \cite{oja-simplified-neuron-model-1982}, a Hebbian learning rule. Other alternatives include more scalable approaches such as  sampling subgraphs that preserve a local structure \citep{mall}, and using specific optimization objectives that approximate the eigenfunctions of the Laplacian~\citep{deng20222neural}.

Recently, \citet{wu2019laplacian} proposed a stochastic approximation to the graph drawing objective, leading to a series of publications on learning options by approximating the eigenfunctions of the Laplacian \cite{jinnai2019discovering,wang2021towards}. An important distinction from \citeauthor{jinnai2020exploration}'s work is that DCEO learns a diverse set of options in parallel, which we have shown to be key for good performance. Arguments for diversity were also presented in work where the options are learned through mutual information-based objectives \cite{gregor16variational,eysenbach2019diversity, CIC}. A shared characteristic between our work and these approaches is the reliance on acting in an off-policy manner. Indeed, each option's experience is leveraged to update any of the agent's components. %This can be a limitation as off-policy algorithms can introduce additional instabilities \citep{sutton2018reinforcement}.
Alternatively, there is a long line of HRL methods derived for the on-policy regime \citep{bacon2017option,harb2018when,ioc}. Future work could be to employ methods that reconcile off-policy and on-policy methods in the HRL setting \citep{NEURIPS2021_24cceab7} or to revise the way options are currently executed.

In RL, there exists a diverse body of works that propose to approximate the eigenfunctions of the graph Laplacian in different ways. \citet{farebrother2023protovalue} propose to leverage the concept of random indicator functions to learn Proto-Value Networks, which are an extension of proto-value functions to the deep learning setting. The authors  show strong performance in the offline RL setting on Atari 2600 games. Closer to the graph drawing objective is the work by \citet{pfau2018spectral}, who introduces Spectral Inference Networks for recovering eigenfunctions of linear operators (and therefore the graph Laplacian). The authors show results in learning interpretable representations from video data. Finally, \citet{ren2023spectral} propose a spectral method that is independent of the policy. More work is required to investigate the properties of each of these approximations. The graph Laplacian and its eigenfunctions knows a variety of use-cases in RL, such as credit assignment \citep{klissarov2020reward}, reward shaping \citep{wu2019laplacian}, and offline representation learning \citep{touati2023does}.

\section{Environments Description}
\label{app:environment_description}
Each of the environments depicted in Figure~\ref{fig:environments} present a different challenge in terms of exploration. \textit{Nine rooms}, a larger variant of the classic Four rooms \citep{sutton1999between}, connects rooms through bottleneck states (hallways). \textit{Maze} is a corridor that turns onto itself and requires persistent exploration for coverage---similar versions of it have been used to highlight the issue of detachment~\cite{ecoffet2021first} some exploration methods face. Finally the \textit{Rubik's cube 2x2} has an underlying topology that is significantly different from traditional grid navigation tasks where all states have the same connectivity and progressing towards the solution state (shown in the middle) requires a discontinuous sequence of actions (i.e. repeating the same action a certain number of times in the cube will bring the agent back to where it started). We use the open source implementation made available by~\citet{kris_cube}.  All environments are stochastic: the agent's action is overwritten by a random action with probability $0.15$.

Episodes are at most $100$ steps long in order to stress the exploration challenge they pose. In each environment, the agent starts from a particular position, shown in blue in Figure \ref{fig:environments}. In the state coverage setting, the agent interacts with the environment until the episode terminates, which happens after $100$ steps. In the reward maximization setting, we add a goal in the bottom left in \textit{Nine rooms}, one goal at each extremity of \textit{Maze}, and one goal for the solution state of the \textit{Rubik's cube 2x2} (located in the middle of Figure \ref{fig:environments}). When the agent reaches the goal it receives a +1 reward (everywhere else is 0) and the episode terminates. Finally, given that the dimensionality of the Rubik's cube's state space is on the order of $10^6$, it is computationally impossible to perform eigendecomposition of an $|\mathscr{S}| \times |\mathscr{S}|$ matrix. For this reason, in the tabular experiments, we restrict the state space to be all the states that are at most three moves away from the solution state. In the experiments with function approximation, none of the tested baselines were able to learn to maximize reward. This is likely due to the fact that the Rubik's Cube is a very challenging domain from the point of view of (1) perception as a single action changes the input features significantly, (2) topology as most of the actions ($\nicefrac{6}{9}$) do not move the agent in a direction towards the solution state. For this reason we also restrict in the function approximation case the total amount of states available to the agent as all states five moves away from the solution state, and we use a tabular algorithm to learn to maximize rewards. Notice that the Laplacian representation is still learned using deep networks which shows the robustness of such an objective.
\looseness=-1

\section{Tabular Option-driven Exploration}\label{sec:results_tabular}
\label{sec:tab_exp}

\subsection{Experimental Setup}
Besides a \textbf{random policy}, we use \textbf{count-based exploration} and $\boldsymbol{\epsilon}$\textbf{z-greedy}~\citep{dabney2021temporally} as baselines. Count-based exploration consists in providing the agent with an intrinsic reward of $\nicefrac{1}{\sqrt{n(s)}}$ at each step, where $n(s)$ is the number of times the agent has visited state $s$. Aside from random exploration, it is likely the most established (and used) algorithm for exploration in RL.
We also consider $\epsilon$z-greedy because it is an option-based exploration algorithm that uses action repetition to obtain temporally-extended exploration, and it has shown significant performance improvements on the Arcade Learning Environment \citep{bellemare2013arcade,machado2018revisiting}. We also considered the Deep Covering Options baseline \citep{jinnai2020exploration}, however we did not include these results as DCEO was significantly outperforming this baseline.

Across all domains, we used a step size of $0.1$ following the experiments by \citet{machado21temporal}. In the state coverage experiments, the agent either takes an action with respect to a random policy or with respect to the exploration approach (CEO, Counts, or $\epsilon$z-greedy). This trade-off was controlled by a hyperparameter which we searched over in the values $\{0.05, 0.1, 0.5, 0.7\}$. For all environments we report the curve for the best performing configuration. The CEO algorithm further introduces two hyperparameters: the size of the option set and the number of options we add/replace at each iteration (this is equal to one episode in our experiments). \citet{machado21temporal} report that an option set size of $1$ and replacing $1$ option per iteration is optimal for the Four rooms domains. We perform a search over the following values $\{1, 3, 5\}$ for each hyperparameter. Preliminary experiments showed that it is more efficient to replace all options at each iteration, therefore we bind the value of the number of options to be replaced with the size of the option set. In the state coverage experiments we found that an option set of size $5$ works best for \textit{Nine rooms} and \textit{Maze}, whereas a size of $1$ was best for \textit{Rubik's cube}. 

In the reward maximization experiments, we leverage each exploration strategy within the Q-learning algorithm \citep{watkins1992q}. The behavior policy is defined to be the $\epsilon$-greedy policy with $\epsilon=0.1$. When the agent chooses an exploratory action it can either execute a primitive action or choose an option. Similarly to the coverage experiments, this trade-off is defined by a hyperparameter that is searched over with the same values as before. We found the option set size of $5$ to be best for all environments. Finally, the option termination for CEO was defined following Theorem 3.1 by \citet{machado2017laplacian}, that is, whenever the agent reaches the (local) maximum value of the option value function.

% {\color{red} We need a whole paragraph here explaining our choices for CEO. How long did we run each phase? How many options did we choose? What were the parameters we used? What were the parameters we used for the other baselines? Did we use $\epsilon$-greedy? How did we terminate the options (this is super important!)? How frequently did we sample an option? How do we even get the information we need about the parameters used by each algorithm?}

\subsection{Results}

\begin{figure*}[t]
     \centering
     \begin{subfigure}[b]{0.3\textwidth}
         \centering
         \includegraphics[width=\textwidth]{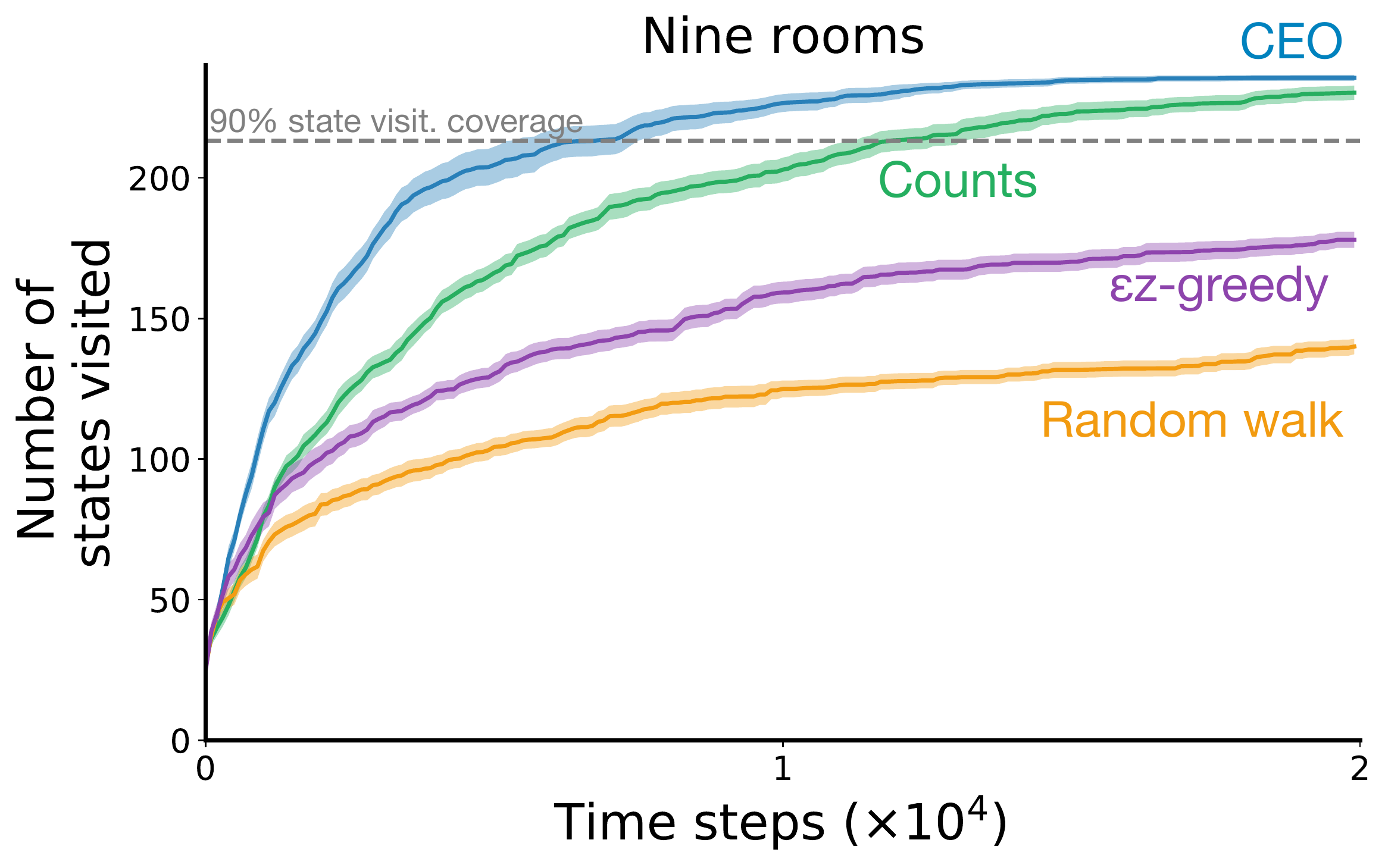}
     \end{subfigure}
     \hfill
     \begin{subfigure}[b]{0.3\textwidth}
         \centering
         \includegraphics[width=\textwidth]{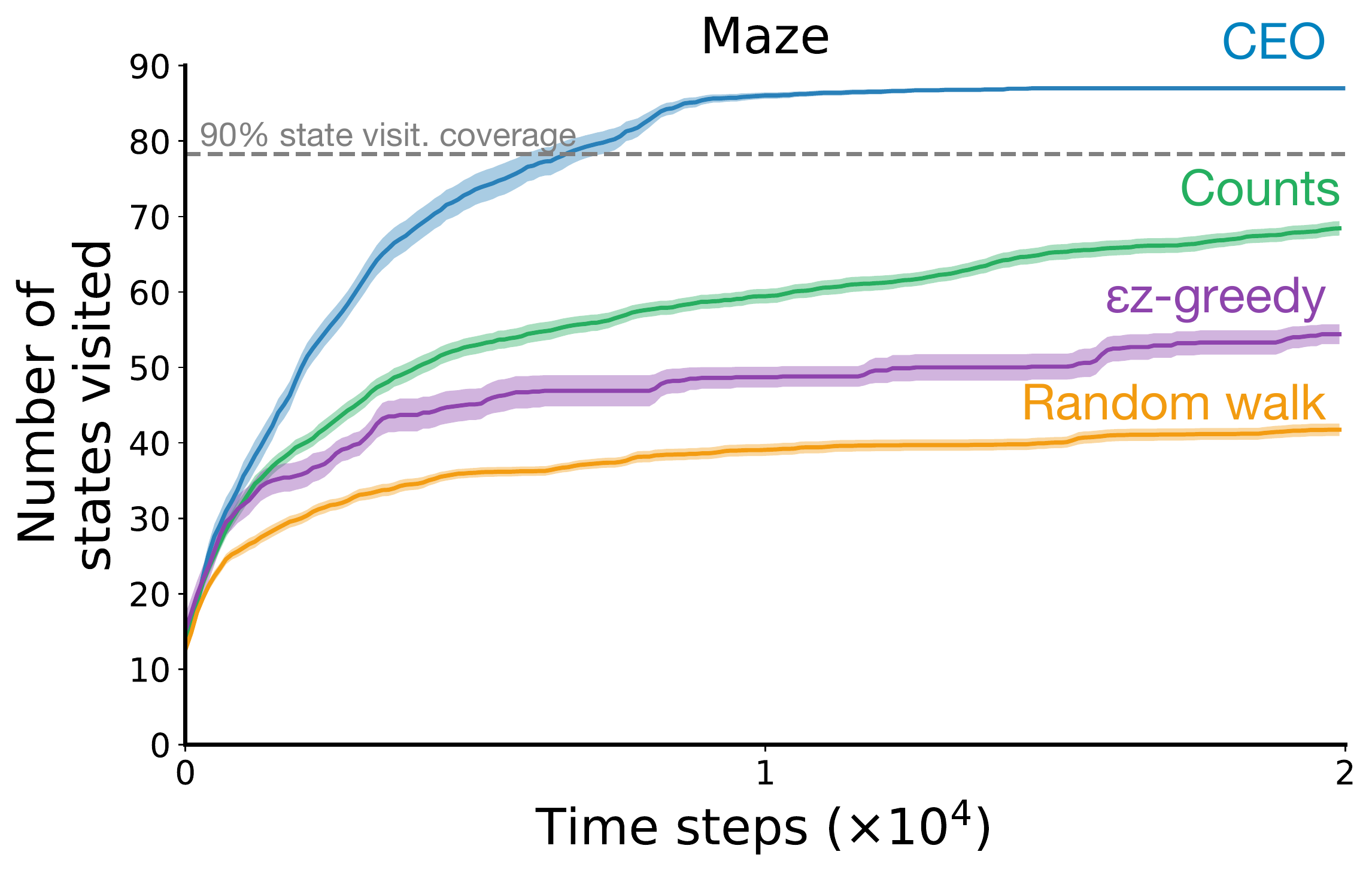}
     \end{subfigure}
     \hfill
     \begin{subfigure}[b]{0.3\textwidth}
         \centering
         \includegraphics[width=\textwidth]{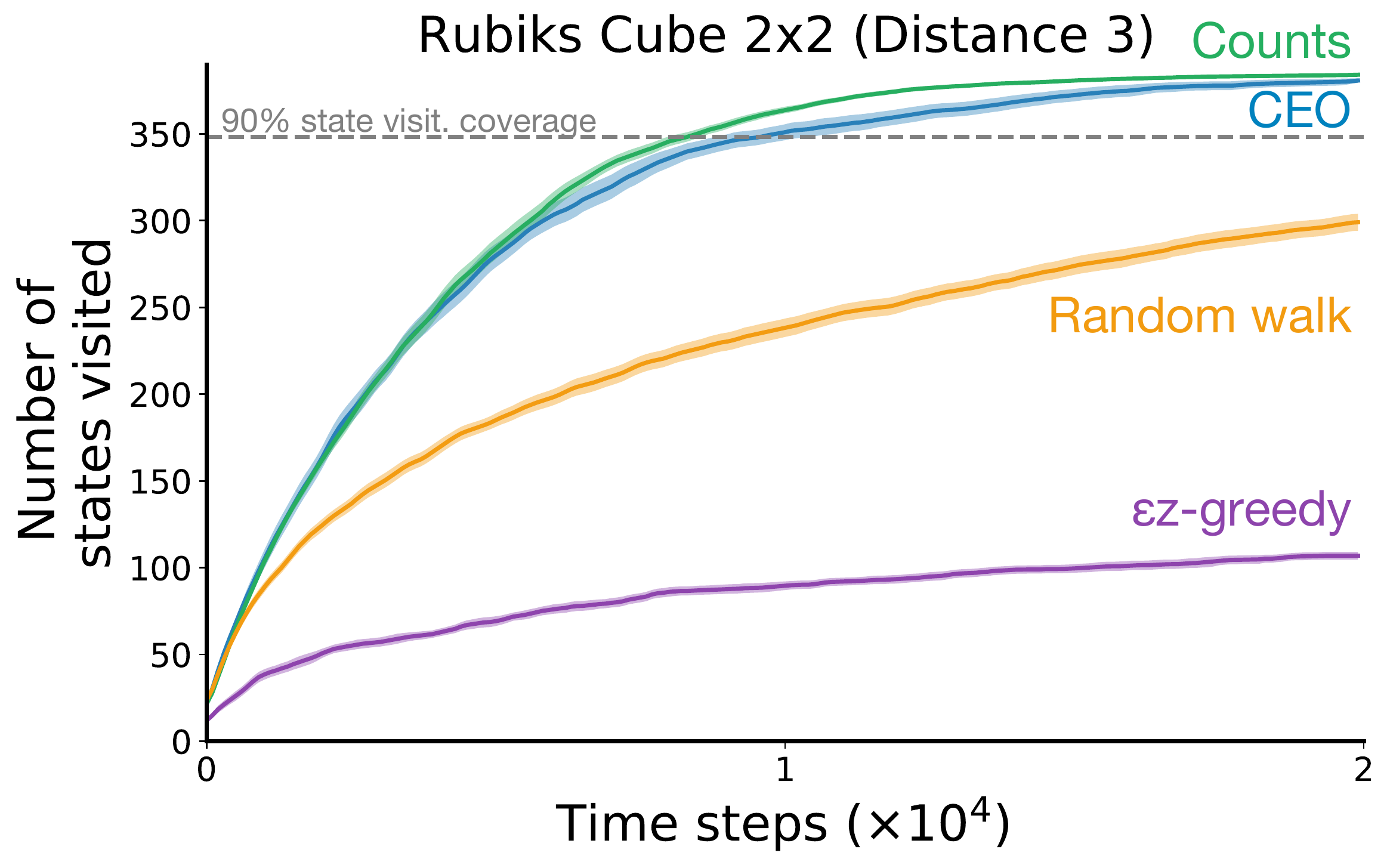}
     \end{subfigure}
        \caption{\textbf{State coverage} in tabular environments. Number of states visited at least once while the agent is acting according to the policy induced by each algorithm. The extrinsic reward is $0$ at every time step. Results show the mean and standard deviation across 30 seeds.}
        \label{fig:tabular_state_coverage}
\end{figure*}

\textbf{State coverage.} We first report results in terms of state coverage in Figure \ref{fig:tabular_state_coverage}. Across all environments, CEO performs either significantly better or is on par with the evaluated baselines. Given the distinct nature of the topology of each environment, this indicates that CEO is robust to different topologies, not being constrained to grid-like navigation tasks.

\begin{figure}[h!]
    \centering
    \includegraphics[width=0.6\columnwidth]{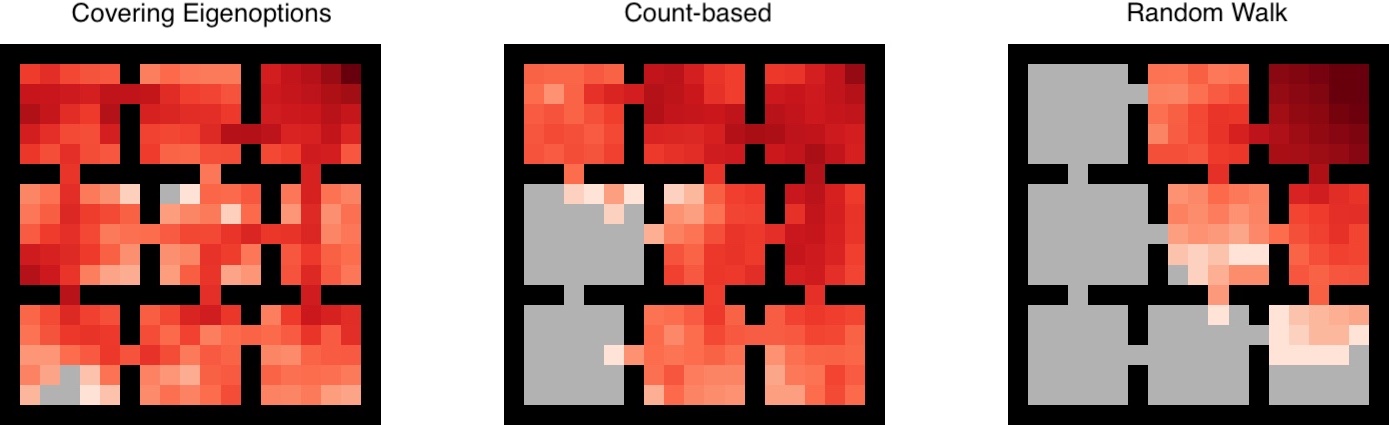}
    \caption{\textbf{State visitation}  after 100 episodes in the Nine rooms domain. State visitation is depicted from white to red (low to high visitation) while gray shows states that were not visited. The paths CEO's options take to efficiently explore the environment leave dark red traces. Count-based exploration is more diffusive in nature and covers states more uniformly from a starting state. }
    \label{fig:state_visitation}
\end{figure}

To better understand CEO's behavior, in Figure \ref{fig:state_visitation} we illustrate the state visitation obtained by some of the baselines after 100 episodes. Darker shades of red represent a higher visitation count for a particular state. We notice that CEO covers the environment in a fundamentally different way when compared to other approaches. While count-based exploration tends to diffuse from a starting state, CEO is much more purposeful, traversing the state space by following the directions defined by the eigenfunctions of the Laplacian, as highlighted by the dark red paths in the left plot of Figure \ref{fig:state_visitation}. This directed nature of exploration is useful in order to cross bottleneck states, as in the \textit{Nine rooms}. It is also especially important in the \textit{Maze} environment as it will incite the agent to reach the limits of the explored state space, thus avoiding the issue of detachment in exploration~\citep{ecoffet2021first}. Notice that it is also in the \textit{Maze} environment that the gap is the greatest between CEO and the baselines. \looseness=-1

\subsection{Reward maximization}

We now evaluate the performance of our method in terms of reward maximization.  In this setting we are faced with a wide range of possibilities as to how the agent may leverage options to maximize reward. In this work, we opt for straightforward solution presented in Algorithm~\ref{algo:ceo}. In particular, we do not learn the value of each option for any given state. Additionally, each option only learns from the intrinsic reward function derived from its associated eigenfunction, making options agnostic to the underlying task. These choices are made such that CEO's improvements in performance may only come from the fact that it leverages a set of options that generate a rich and diverse stream of experience for learning.

\begin{figure*}[h!]
     \centering
     \begin{subfigure}[b]{0.3\textwidth}
         \centering
         \includegraphics[width=\textwidth]{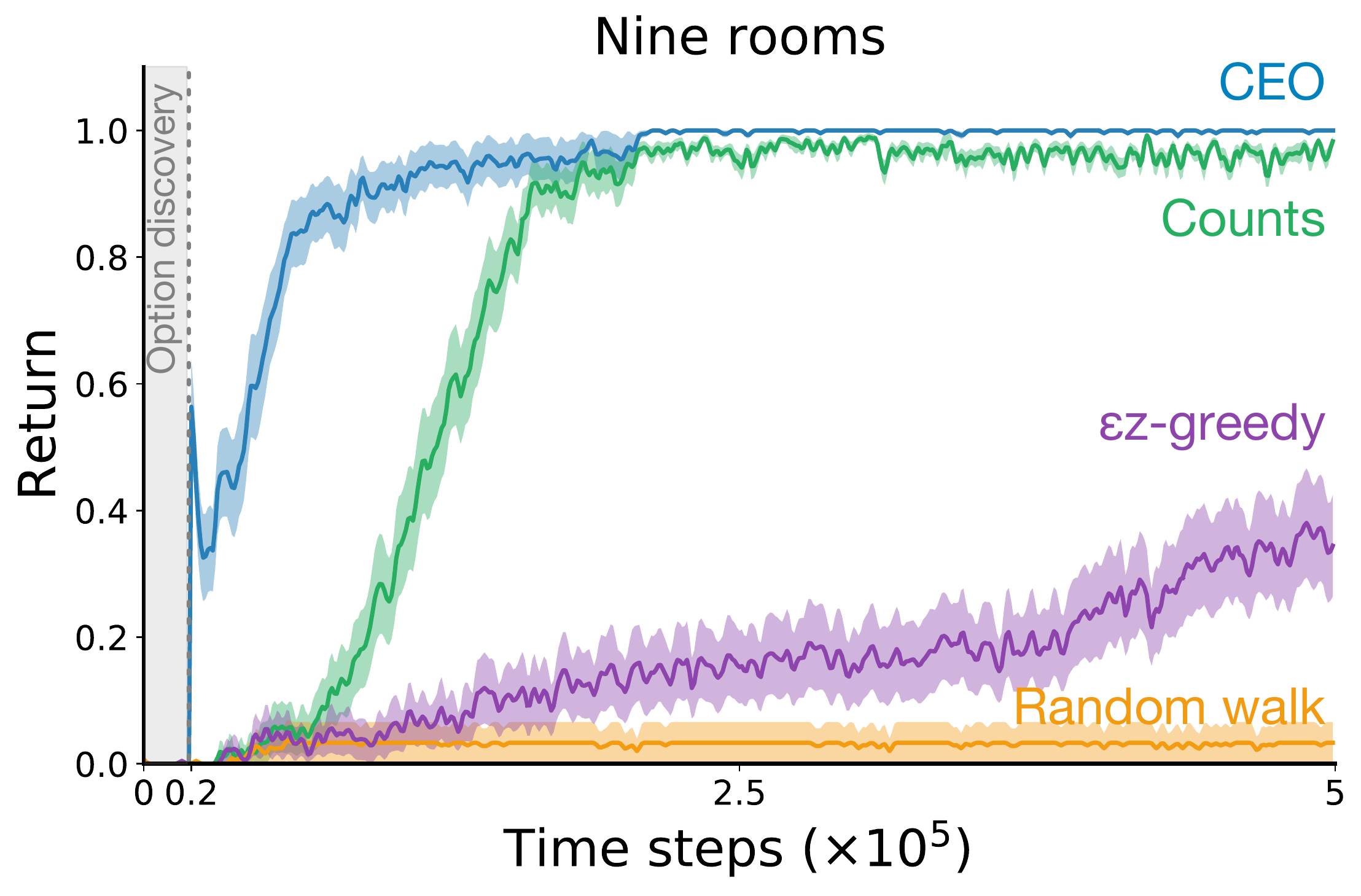}
     \end{subfigure}
     \hfill
     \begin{subfigure}[b]{0.3\textwidth}
         \centering
         \includegraphics[width=\textwidth]{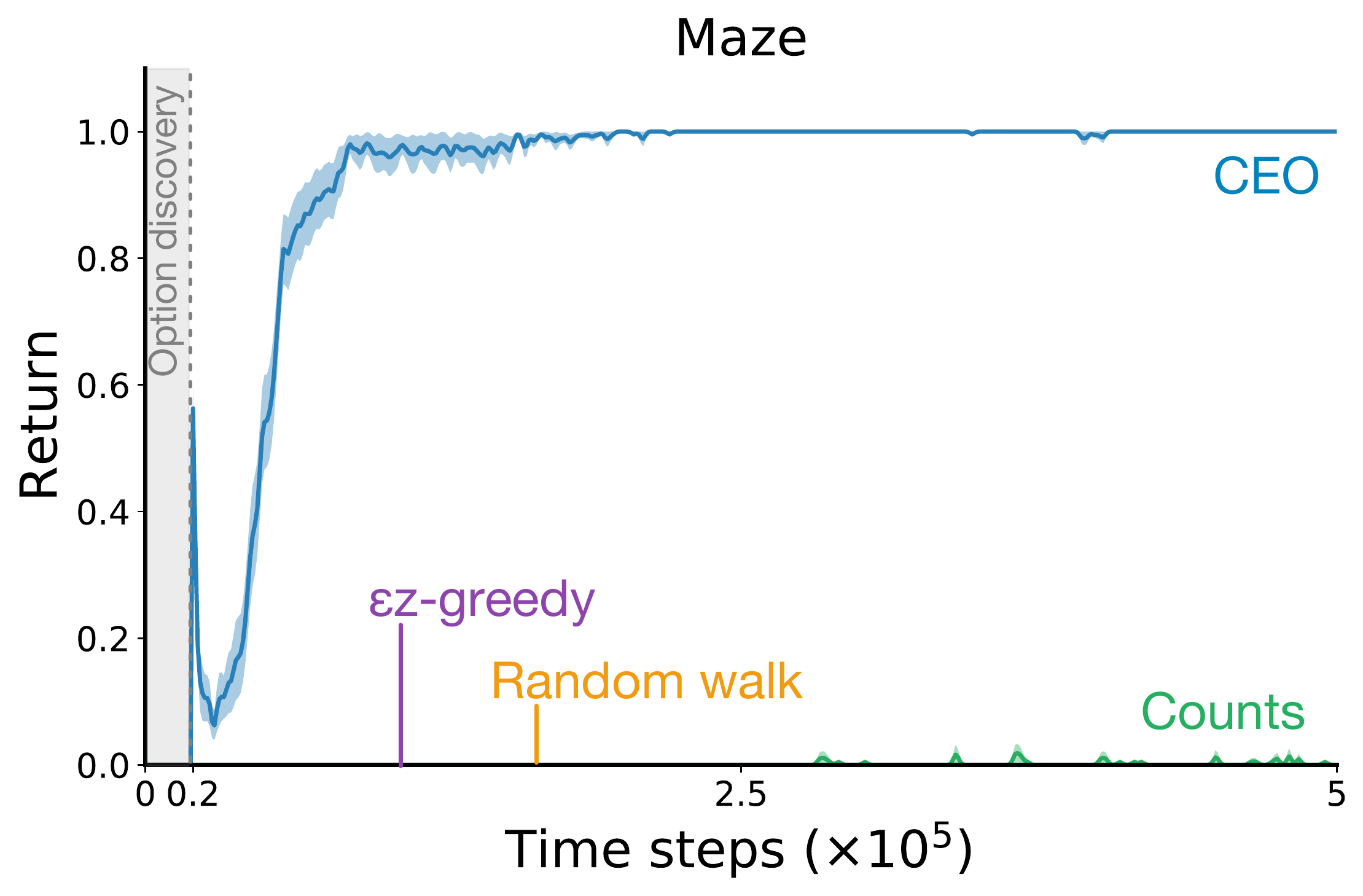}
     \end{subfigure}
     \hfill
     \begin{subfigure}[b]{0.3\textwidth}
         \centering
         \includegraphics[width=\textwidth]{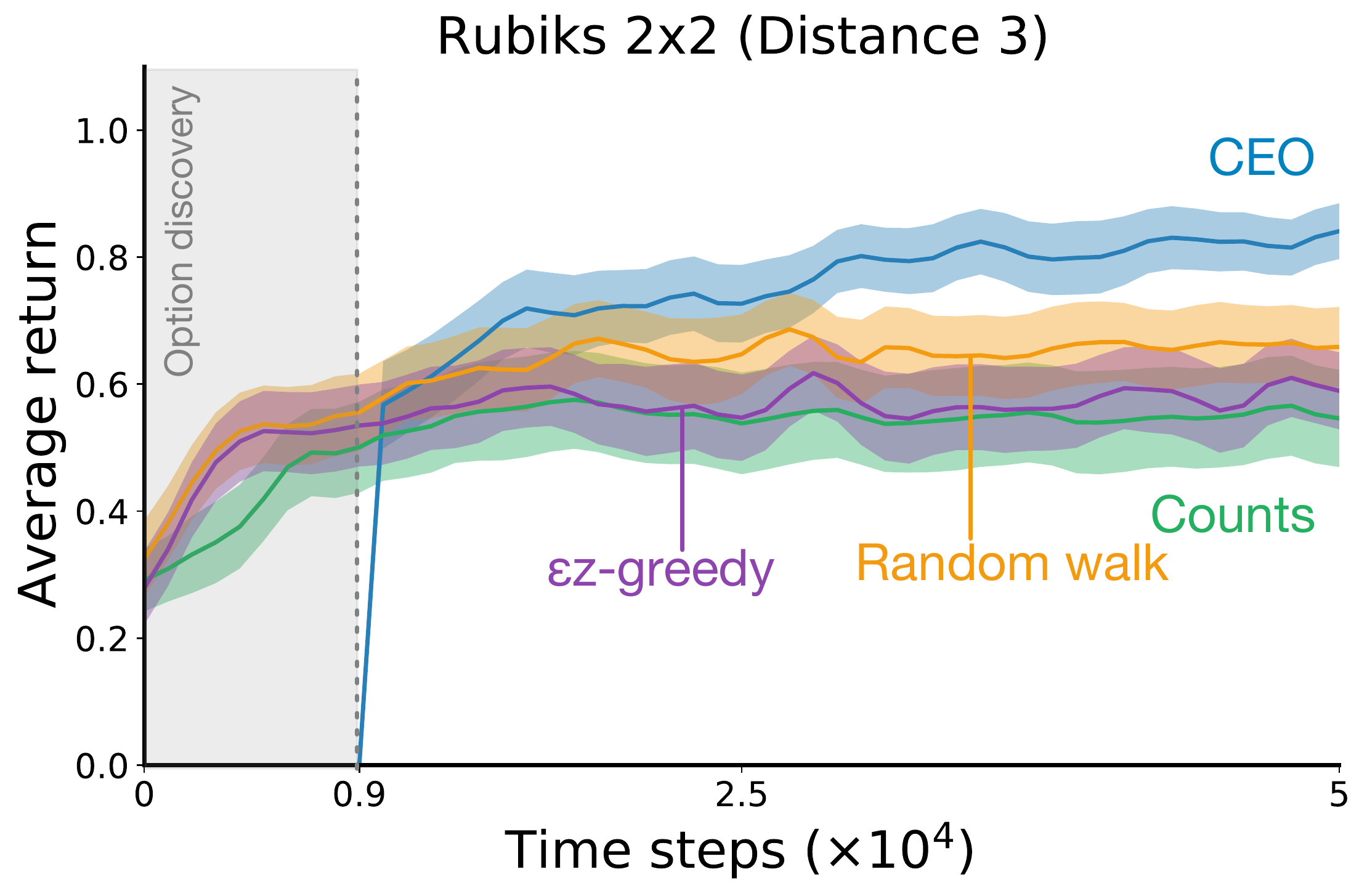}
     \end{subfigure}
        \caption{\textbf{Reward maximization} in tabular environments. CEO uses a two-phased algorithm where it first pretrains a set of options through intrinsic motivation before leveraging them for reward maximization. CEO's curve is delayed by the amount of time spent in the first phase of option discovery. Despite this additional cost, we notice that it produces strong performance across all domains. When the curves of different methods are not visible inside the shaded region it is because the exploration strategy used by the baseline method did not lead to a single positive reward during the whole period. Results show the mean and standard deviation across 30 seeds.}
        \label{fig:tabular_return_maximization}
\end{figure*}

The empirical setup is the same as in the state coverage experiments, except for the addition of a goal which gives +1 reward while all other states give 0. We provide a detailed description in Appendix~\ref{app:environment_description}. For all environments, we first learn a set of five options in a reward-free setting for $T_{discovery}$ steps before fixing them. These options are then executed by the CEO agent in order to explore the environment and learn a reward maximizing policy. To account for the time CEO spent in the option discovery phase (shaded region in Figure~\ref{fig:tabular_return_maximization}), we delay the start of CEO's performance curve by for $T_{discovery}$ steps. Figure \ref{fig:tabular_return_maximization} shows that, even when considering the delay induced by the option discovery phase, CEO is not only able to learn faster than other methods but it also learns better policies across all environments.

\begin{algorithm}[h!]
\caption{Two-phased CEO Algorithm}
\begin{algorithmic}[1]
% \STATE $D \leftarrow 10$ \# Average option duration
\STATE $\epsilon \leftarrow 1.0$
\STATE Reset environment and observe state $s$\STATE $\bot = \textrm{True}$ \# \textit{Episode termination}
\FOR{$i = 1$ {\bfseries to} $T$}
\IF{$\bot$}
\STATE $\tau \leftarrow \textrm{True}$ \# \textit{Option termination}
\STATE $o \leftarrow -1$ \# \textit{No active option}
\ENDIF
\STATE $\tau \leftarrow U(0,1) < \nicefrac{1}{D} \text{ } \vee  \text{ } \tau$
\IF{$\tau$}
\IF{$U(0, 1) < \epsilon$}
\IF{$U(0,1) < \mu$}
\STATE $o \sim U(|\mathscr{O}|)$
\STATE $\tau \leftarrow \textrm{False}$
\STATE $a \sim \pi_o(\cdot|s)$
\ELSE
\STATE $o \leftarrow -1$
\STATE $\tau \leftarrow \textrm{True}$
\STATE $a \sim U(\mathscr{A})$
\ENDIF
\ELSE
\STATE $a \leftarrow \max_{a \in \mathscr{A}} Q(s,a)$
\ENDIF
\ELSE
\STATE $a \sim \pi_o(\cdot|s)$
\ENDIF
\STATE Execute action $a$, observe $r, s', \bot$
\STATE Store transition $(s,a,r,s')$ in buffer $B$
\STATE Sample a minibatch of transitions $(s_j, a_j, r_{j+1}, s_{j+1})$
\IF{$t < T_{discovery} \land \bot$}
\STATE $\forall o_i \in \mathscr{O}, r_j \leftarrow f_i(s') - f_i(s)$
\STATE Perform eigendecomposition to obtain the eigenfunctions
\STATE Train each option until convergence using its intrinsic reward
\ELSE
\STATE $\epsilon \leftarrow 0.1$
\STATE Train main learner on extrinsic reward using Q-Learning
\ENDIF
\STATE $s \leftarrow s'$
\ENDFOR
\end{algorithmic}
\label{algo:ceo}
\end{algorithm}

\clearpage

\section{Coverage Results Using Deep Neural Networks}
\label{app:coverage_fa}
\begin{figure*}[h!]
     \centering
     \begin{subfigure}[b]{0.3\textwidth}
         \centering
         \includegraphics[width=\textwidth]{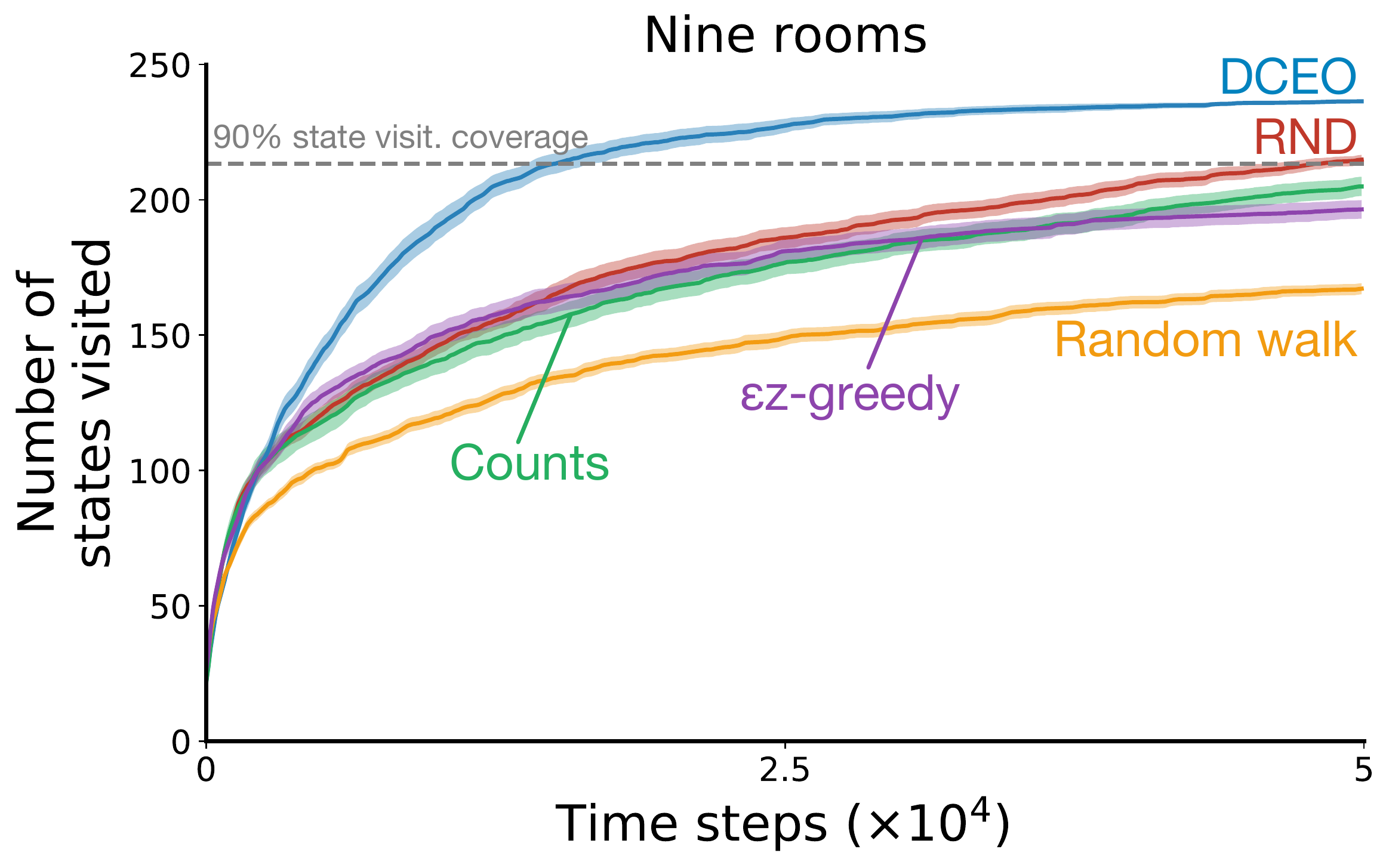}
     \end{subfigure}
     \hfill
     \begin{subfigure}[b]{0.3\textwidth}
         \centering
         \includegraphics[width=\textwidth]{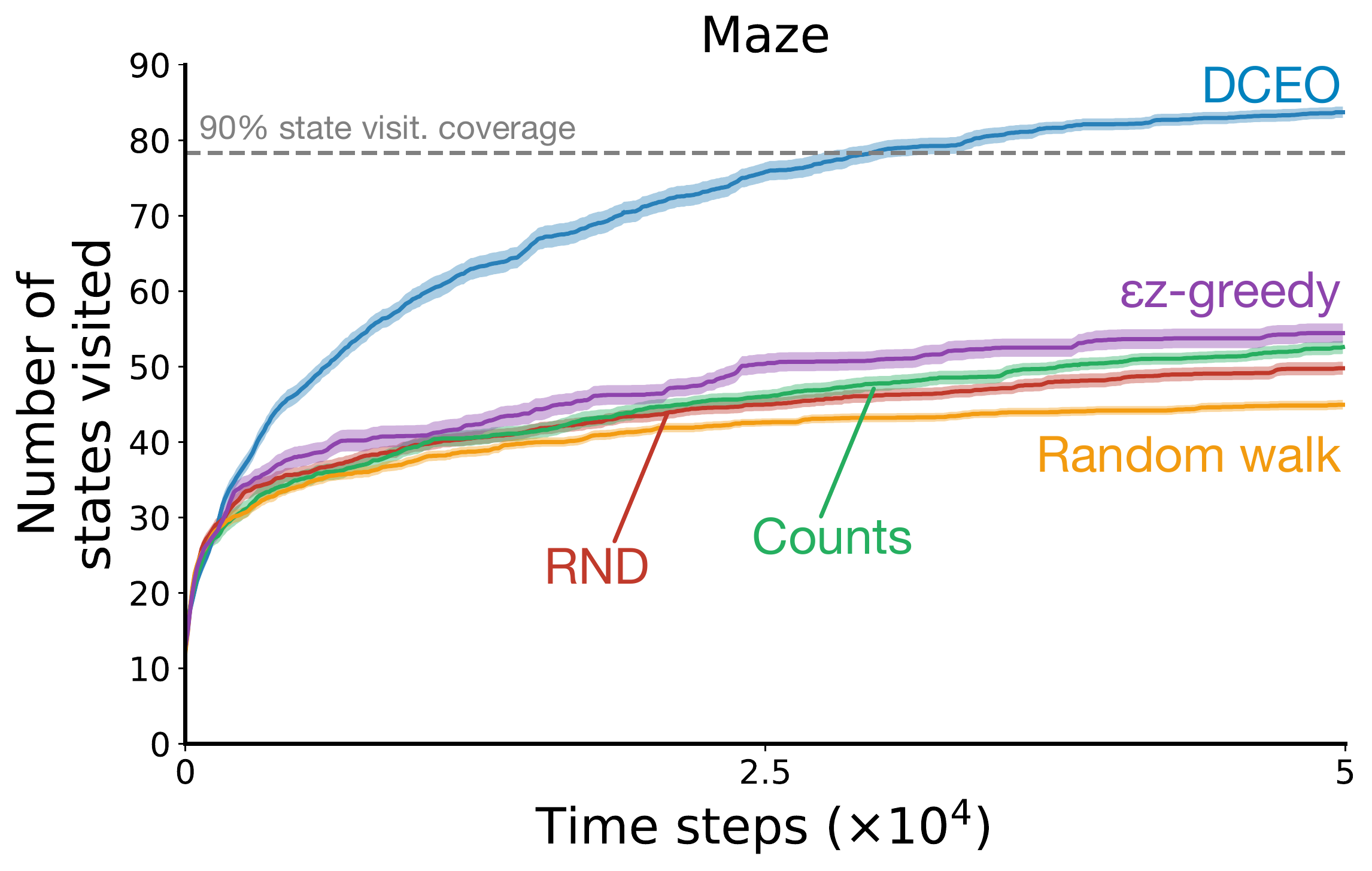}
     \end{subfigure}
     \hfill
     \begin{subfigure}[b]{0.3\textwidth}
         \centering
         \includegraphics[width=\textwidth]{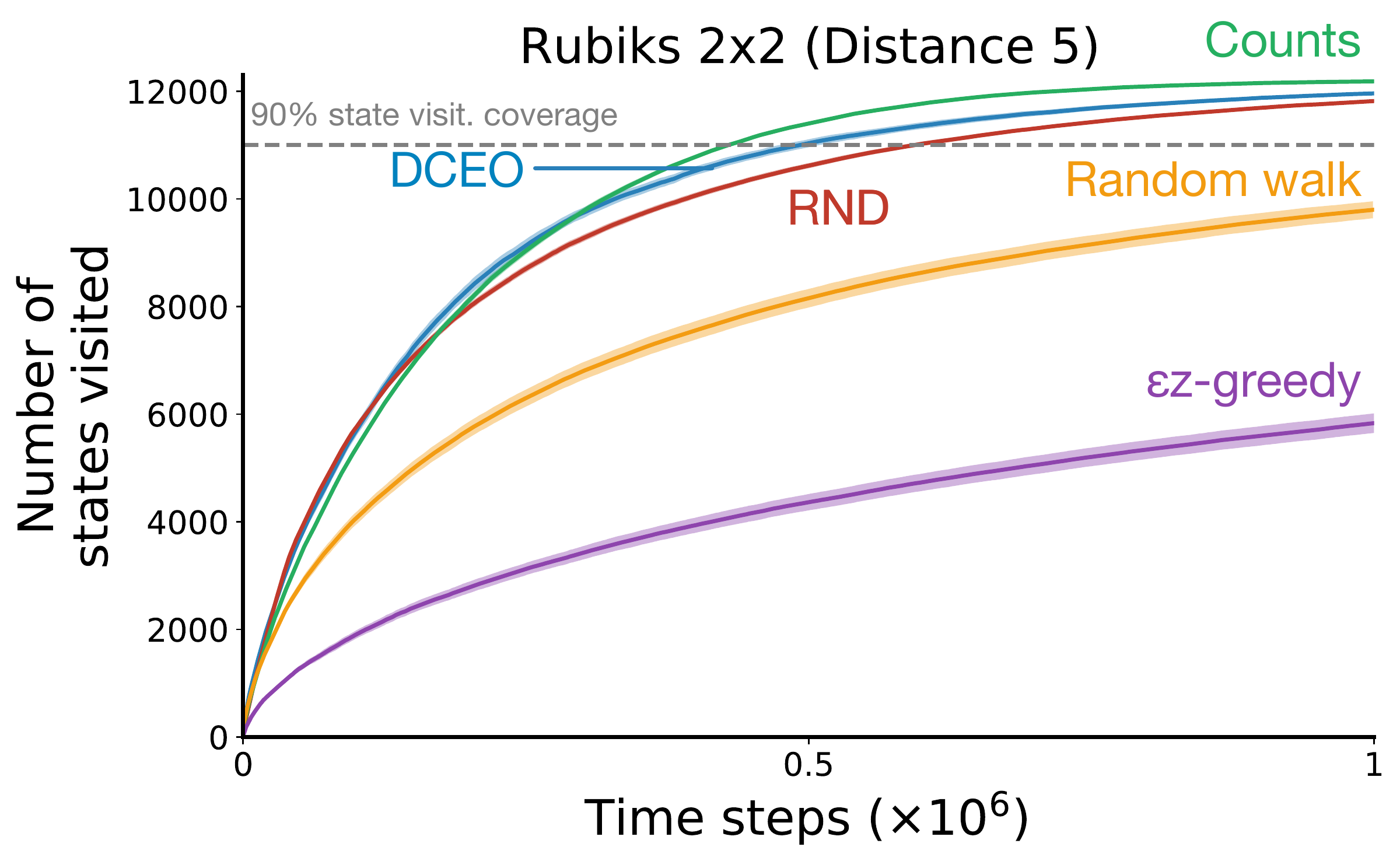}
     \end{subfigure}
        \caption{\textbf{State coverage} under deep function approximation. We show the number of states visited at least once while the agent acts according to the policy induced by each algorithm. The environment reward is $0$ at every time step. Results show the mean and standard deviation across 30 seeds.}
        \label{fig:fa_coverage_maximization}
\end{figure*}

We present the results for state coverage using the DCEO algorithm in Figure \ref{fig:fa_coverage_maximization}. We notice that DCEO is either significantly more efficient or on-par at covering the state space.

\section{Comparison Between Different Laplacian Objectives}

\begin{figure}[h!]
     \centering
     \begin{subfigure}[b]{0.33\columnwidth}
         \centering
         \includegraphics[width=\columnwidth]{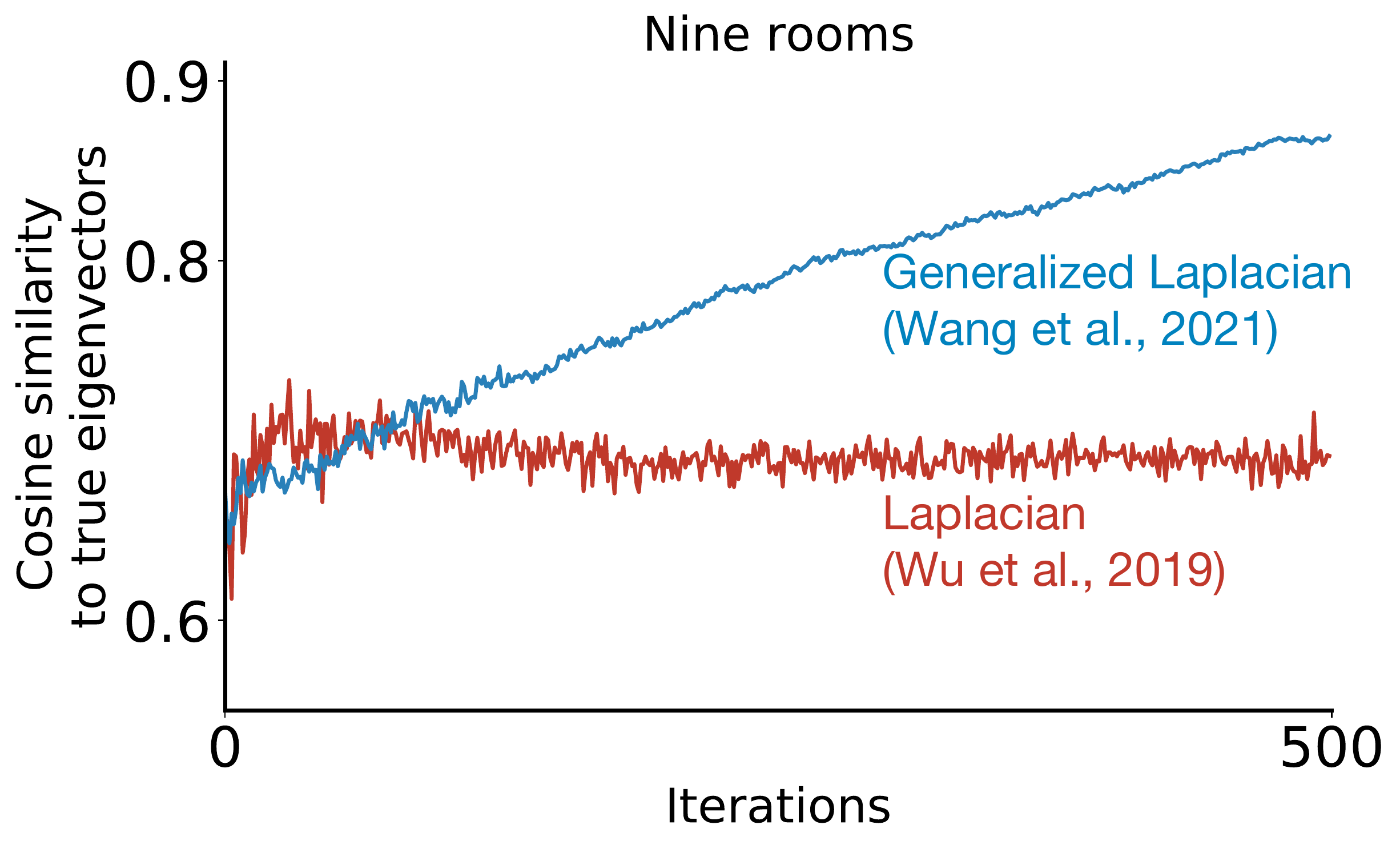}
     \end{subfigure}
    %  \hfill
     \begin{subfigure}[b]{0.33\columnwidth}
         \centering
         \includegraphics[width=\columnwidth]{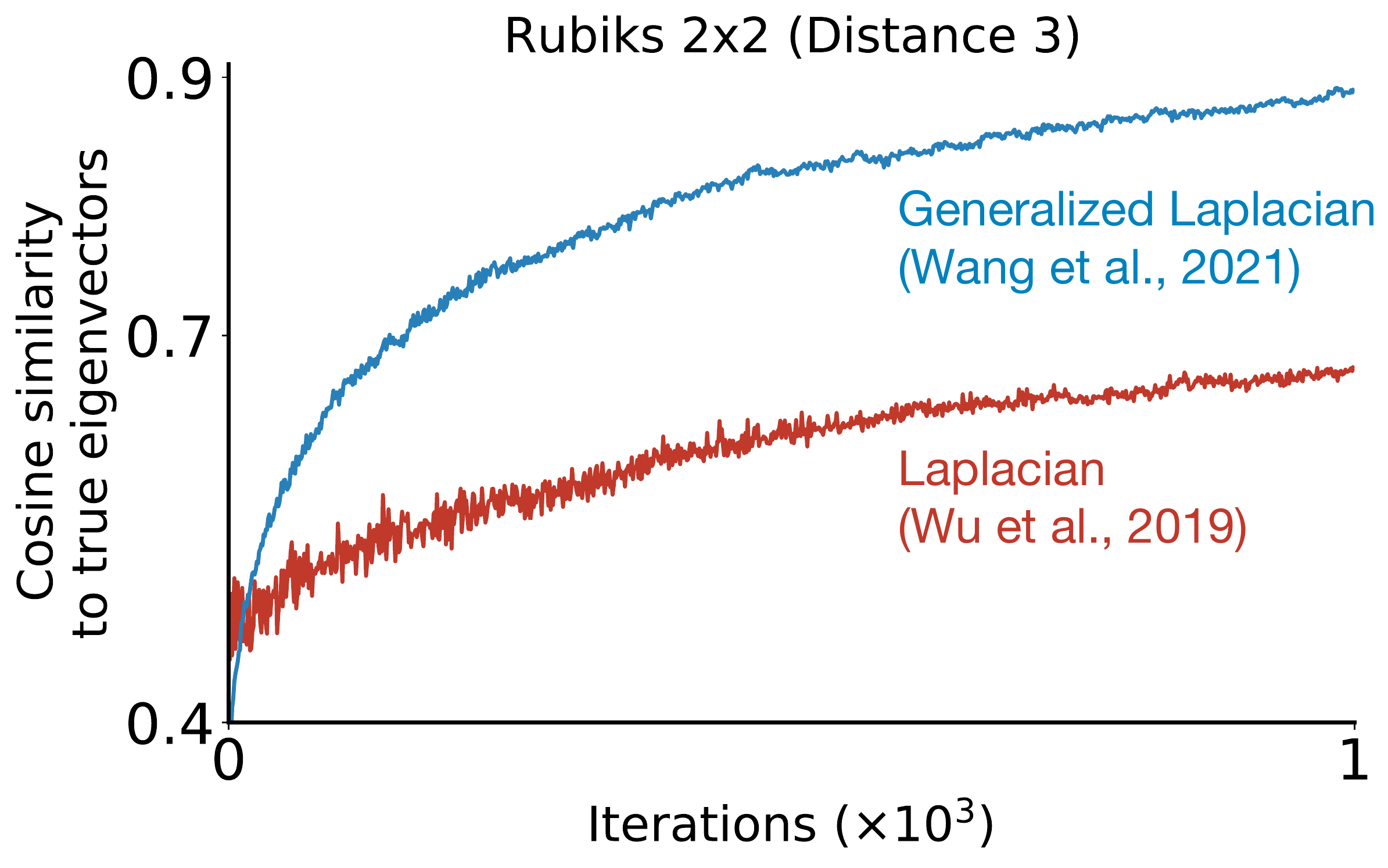}
     \end{subfigure}
        \caption{Comparison between Laplacian (Wu et al., 2019) and Generalized Laplacian (Wang et al., 2021).}
        \label{fig:compare_obj}
\end{figure}

In our experience, maximizing the generalized Laplacian instead of the original objective proposed \citet{wu2019laplacian} was crucial for obtaining the performance we report. We illustrate this point in Figure \ref{fig:compare_obj} where only the generalized Laplacian objective recovers accurate approximations to the eigenfunctions of the Laplacian.

\section{Comparison Between the Fully Online DCEO algorithm and DCEO with phases}
\label{app:compare_dceo}

\begin{figure}[h!]
     \centering
     \begin{subfigure}[b]{0.33\columnwidth}
         \centering
         \includegraphics[width=\columnwidth]{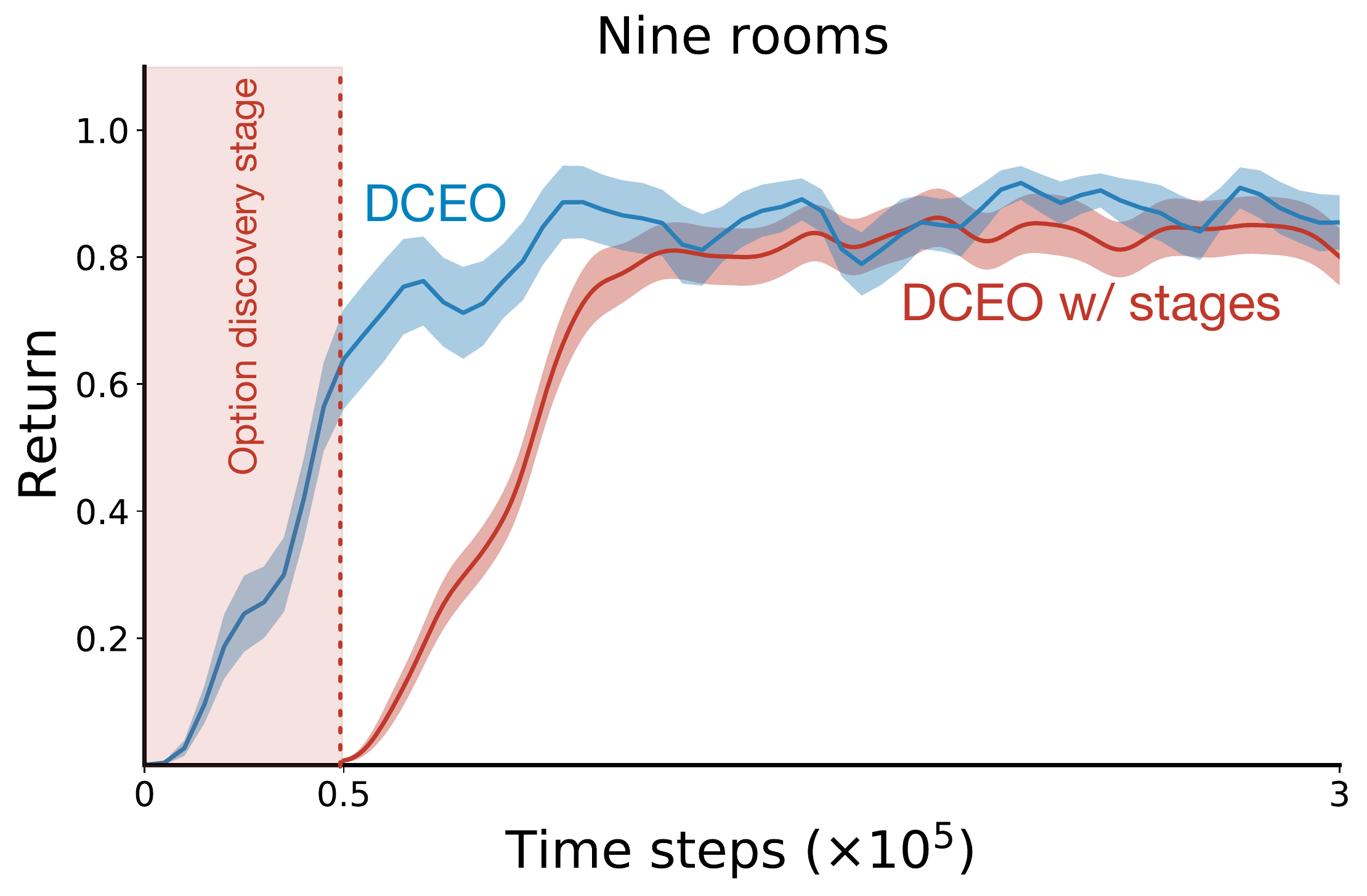}
     \end{subfigure}
    %  \hfill
     \begin{subfigure}[b]{0.33\columnwidth}
         \centering
         \includegraphics[width=\columnwidth]{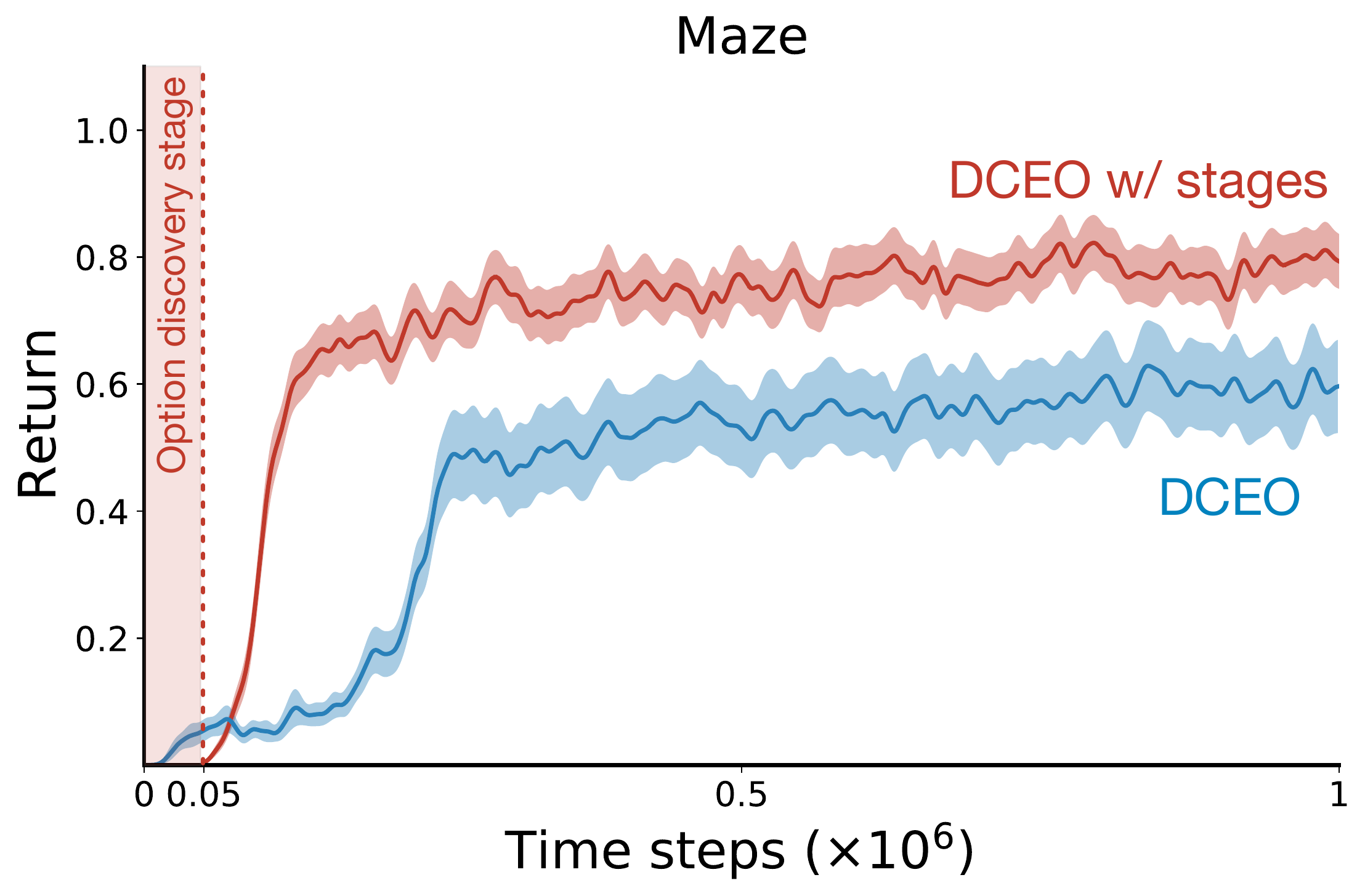}
     \end{subfigure}
        \caption{Comparison between learning DCEO fully online vs the staged one.}
        \label{fig:comparison_online_vs_stages}
\end{figure}

In Figure \ref{fig:comparison_online_vs_stages} we compare the performance of the fully online DCEO algorithm (shown simply as DCEO) with the two-phased version of DCEO (shown as \textit{DCEO w/ stages}) that has access to a pretraining phase for option discovery. We notice that the fully online DCEO is a strong algorithm that performs relatively well when compared to the two-phased version. This online version removes the additional complexities of having a pretraining phase, is more generally applicable and scales naturally.

\section{Visualizations in Escape Room}
\label{app:visual_escape}

\begin{figure}[h!]
     \centering
     \begin{subfigure}[b]{.8\columnwidth}
         \centering
         \includegraphics[width=\columnwidth]{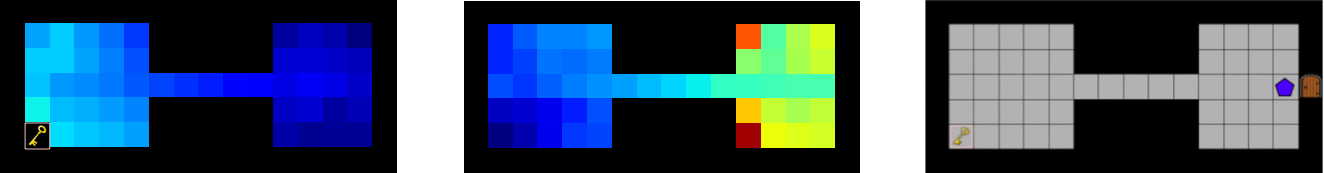}
     \end{subfigure}
        \caption{Visualization of the first eigenfunction in Escape Room. On the left we show its values for all agent positions before picking up the key. In the middle we show its values for all agent positions after picking the key. On the right we show the environment. }
        \label{fig:eigen_visualization}
\end{figure}

We present a visualization of the first eigenfunction of the generalized Laplacian in Figure \ref{fig:eigen_visualization} and we show its value for each agent position before picking up the key (figure on the left), as well as after picking up the key (figure on the middle). We witness an interesting property: the first eigenfunction, which encodes the principal direction in the environment, seeks to pick the key and traverse, in the opposite way, the state space.
In this environment, when the agent picks up the key, the observation features change in a consistent way and the agent is led to a whole new part in the feature space. This important transition is naturally encoded in the Laplacian representation and helps the agent to effectively explore its environment. Picking up the key therefore could be seen as traversing a bottleneck state.

\section{Non-Stationary Environments}
\label{app:non_stationary}

\begin{figure}[h!]
     \centering
     \begin{subfigure}[b]{0.4\columnwidth}
         \centering
         \includegraphics[width=\columnwidth]{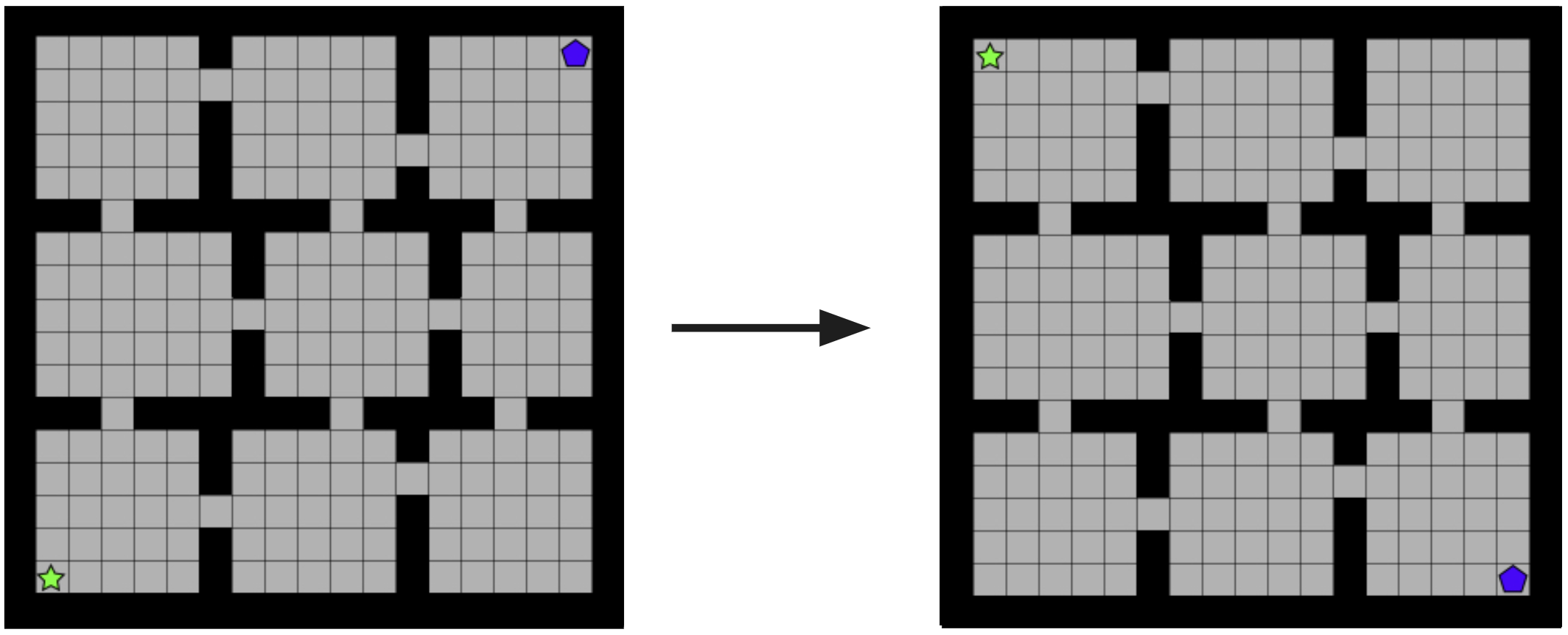}
         \caption{Changing the goal position}
     \end{subfigure}
     \hspace{1cm}
     \begin{subfigure}[b]{0.4\columnwidth}
         \centering
         \includegraphics[width=\columnwidth]{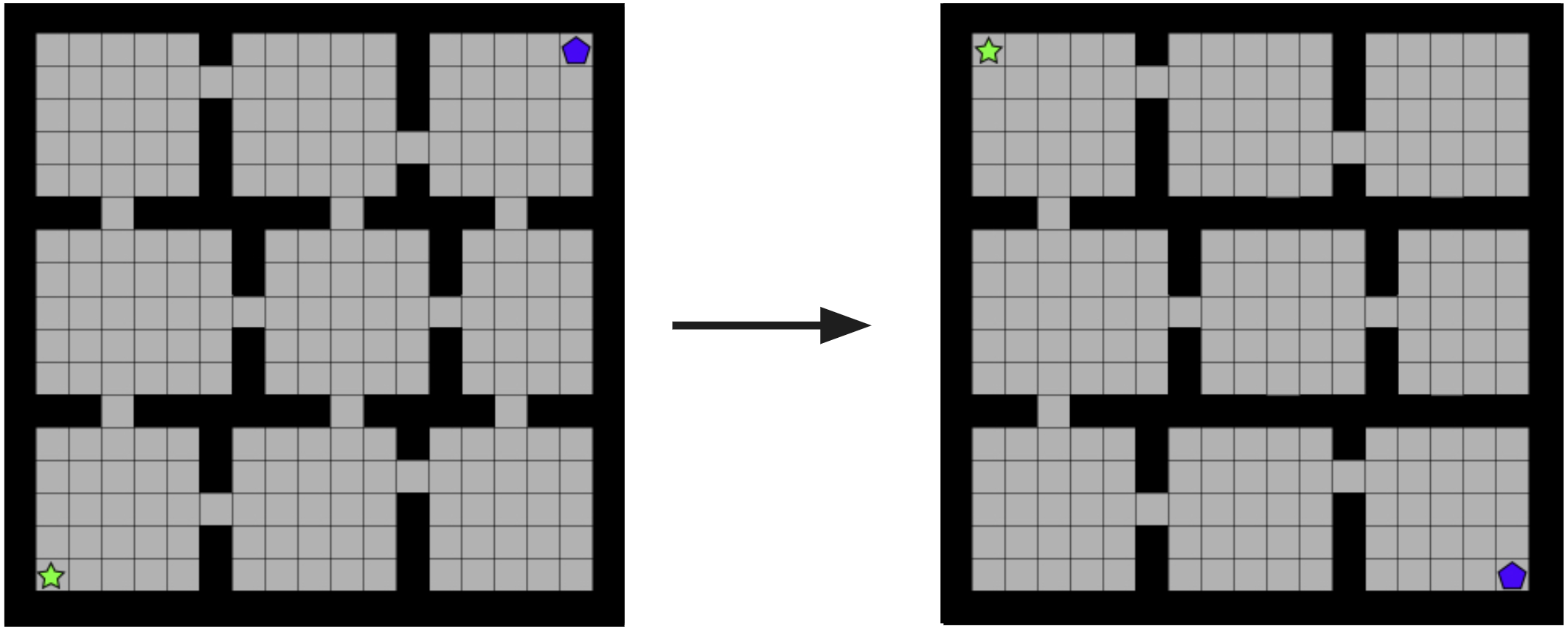}
         \caption{Changing the topology and goal}
     \end{subfigure}
        \caption{Visualization of the non-stationary tasks evaluated in Section~\ref{sec:transfer}. In the first experiment, both the location of the agent and the goal are changed. The agent does not get to observe where the new goal location is (it is invisible in input space). In the second experiment, the location of the agent and the goal are changed, and the topology of the environment is altered as well. Some of the bottleneck states are closed off, forcing the agent to find a different path to the goal.}
        \label{fig:non_stationary_envs}
\end{figure}

\section{Experimental Details}
\label{app:experimental_details}
RND and count-based exploration produce a bonus for exploration that is added to the environment reward. This bonus is multiplied by a hyperparameter $\beta$, for which we searched over the following values: $\{ 0.25, 0.5, 0.75, 1.0 \}$ for RND and $\{0.0001, 0.001, 0.01, 0.1, 1.0\}$ for Counts. We report the best performing curve for each environment. For $\epsilon$z-greedy the parameter for the $\zeta$ distribution was taken to be $k=2$ following best results by \citet{dabney2021temporally}. On initial experiments, different values of $k$ did not lead to significantly better performance.
In the case of the other option-based algorithms (DIAYN, $\epsilon$z-greedy, and DCEO), we execute options with a certain probability $\mu$ whenever the DDQN algorithm is not acting greedily (i.e. when $U(0,1) < \epsilon$). For each algorithm we searched over values in $\{0.2, 0.7, 0.9\}$ or $\{0.2, 0.7, 0.8\}$. Additionally, when leveraging options we must define the size of the option set $N$, which was searched over in $\{3, 5, 10\}$. Across all environments an option set of $10$ performed well, except for Rubik's Cube where a value of $3$ was better. For the values of $\mu$ for DCEO, $\epsilon$z-greedy, and DIAYN (where applicable): in Figure \ref{fig:fa_return_maximization} from left to right, we used $0.9, 0.9, 0.7$; in Figure \ref{fig:fa_return_maximization_online} from left to right, we used $0.8$; in Figure \ref{fig:non_stationarity} we used $0.7, 0.8$; in Figure \ref{fig:fancy_fa_return_maximization} we used $0.9, 0.2, 0.7$. It is important to note that the algorithm performed generally well for high values of $\mu$. However, since we also tuned all the other baselines for each environment, we present the best performing choice of hyperparameters for all methods. An important detail of the count-based baseline is that we do not estimate pseudo-counts \citep{bellemare2016unifying}, we instead provide the agent with \textit{perfectly accurate state counts}, which provide a significant advantage to this baseline. The results present the mean and standard deviation obtained across 30 independent seeds.\\

For all deep learning experiments we used a step size of $0.0001$. This was chosen after a first investigation over the range $\{ 0.001, 0.0005, 0.0003, 0.0001, 0.00005 \}$ on the simpler environments (Nine rooms and Maze). 

The networks used were the following. The convolutional networks were a two layered convolutional network of channels 32, kernel 3 by 3 and stride 2. This was followed by a fully connected layer of size 256, followed by the outputs of the networks. All activations were ReLUs. For the Rubik's experiments, we used a stacked of 3 fully connected layers of size 256 before the outputs. All activations were ReLUs.

In the high dimensional experiments from Section \ref{scale_up}, we simply use the best performing hyperparameter configuration from previous experiments to obtain results for DCEO, that is the option selection probabiliy $\mu$ is $0.9$ and the option duration $D$ is $10$. The number of options is reduced to $5$ to allow for a faster algorithm iteration. The networks used is the standard Nature DQN convolutional network for the 3D Navigation experiments whereas we use the standard Rainbow network for Atari.

\section{Visualizing the Eigenfunction Through Time}
\label{app:additional_visuals}
In this section we present visualizations of the first 10 dimensions of the Laplacian representation as the agent learns to reach the goal. We present such visualizations on the Nine rooms environments which we previously described and used as a benchmark to compare DCEO to other algorithms. Additionally, we present results on the GridMaze environment which was previously used by \citet{wang2021towards}. We do so to contrast the way the Laplacian representation evolves and converges in our setting, where the agent has to discover the state space (as each episode starts from the bottom left)  and where it seeks to reach a goal state, which was not the case of previous work.

\begin{figure}[h!]
    \centering
    \includegraphics[width=1.\textwidth]{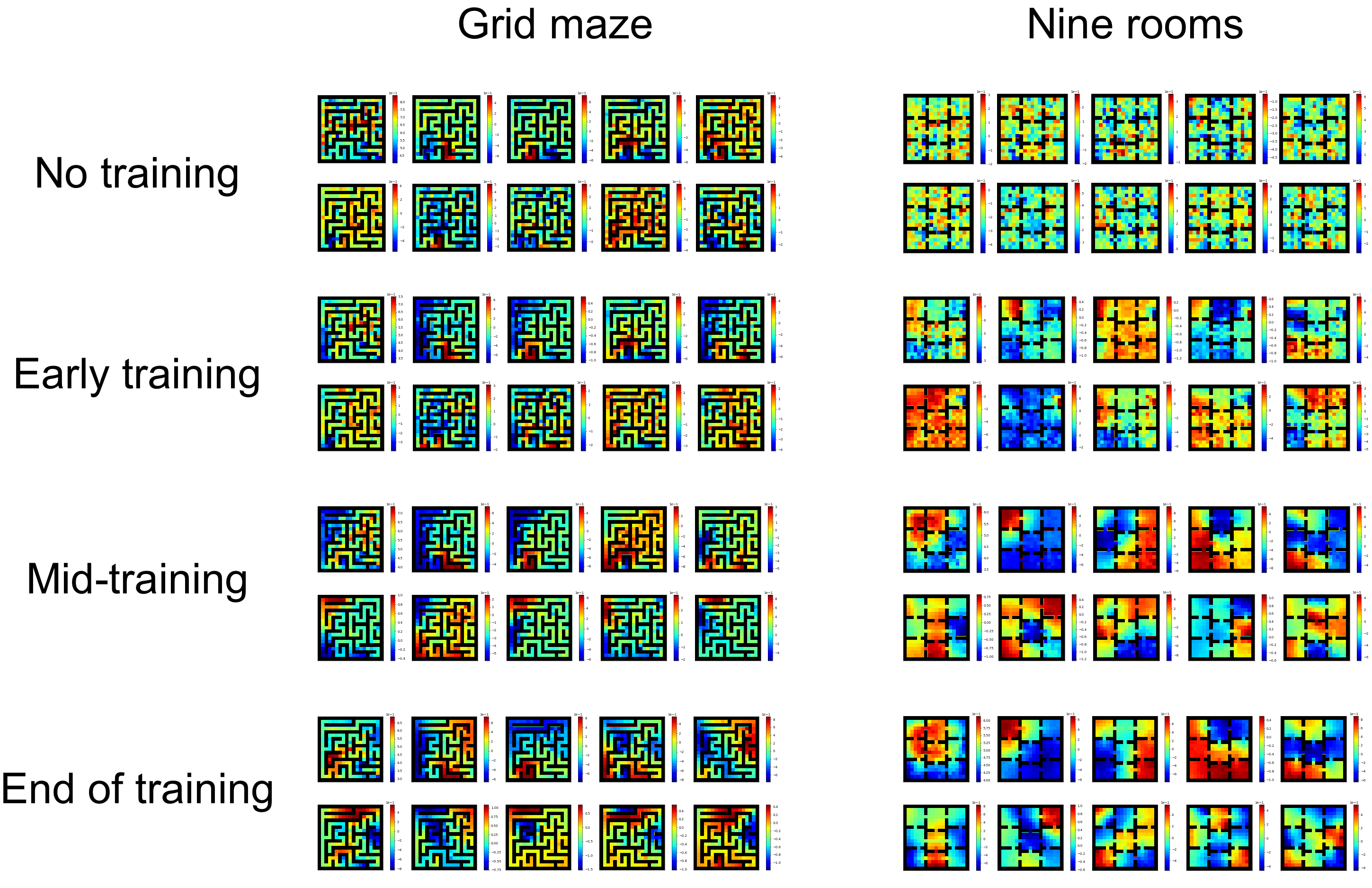}
    \caption{Visualization of the first 10 dimensions of the Laplacian presentation (the first 10 eigenfunctions) learned by the DCEO agent as learning progresses. We notice that at first, all representations are random. As the agent covers more states, the objective's orthogonality constraint leads to representations that point in different directions on a subset of the state space. Finally, as the agent has discovered the goal and learns to maximize for it, the representations take the agent's policy into consideration as well as the general topology of the environment. This is in contrast to the representaion found by previous work \citep{wang2021towards} where the agent can respawn anywhere in the grid and do not seek to maximize a reward.}
    \label{fig:extra_visuals}
\end{figure}

\clearpage
\section{Deep Covering Eigenoptions: A Two-Phased Version}
\label{app:algo}

\begin{algorithm}[h!]
\caption{Two-phased DCEO Algorithm}
\begin{algorithmic}[1]
% \STATE $D \leftarrow 10$ \# Average option duration
\STATE $\epsilon \leftarrow 1.0$
\STATE Reset environment and observe state $s$\STATE $\bot = \textrm{True}$ \# \textit{Episode termination}
\FOR{$i = 1$ {\bfseries to} $T$}
\IF{$\bot$}
\STATE $\tau \leftarrow \textrm{True}$ \# \textit{Option termination}
\STATE $o \leftarrow -1$ \# \textit{No active option}
\ENDIF
\STATE $\tau \leftarrow U(0,1) < \nicefrac{1}{D} \text{ } \vee  \text{ } \tau$
\IF{$\tau$}
\IF{$U(0, 1) < \epsilon$}
\IF{$U(0,1) < \mu$}
\STATE $o \sim U(|\mathscr{O}|)$
\STATE $\tau \leftarrow \textrm{False}$
\STATE $a \sim \pi_o(\cdot|s)$
\ELSE
\STATE $o \leftarrow -1$
\STATE $\tau \leftarrow \textrm{True}$
\STATE $a \sim U(\mathscr{A})$
\ENDIF
\ELSE
\STATE $a \leftarrow \max_{a \in \mathscr{A}} Q(s,a)$
\ENDIF
\ELSE
\STATE $a \sim \pi_o(\cdot|s)$
\ENDIF
\STATE Execute action $a$, observe $r, s', \bot$
\STATE Store transition $(s,a,r,s')$ in buffer $B$
\STATE Sample a minibatch of transitions $(s_j, a_j, r_{j+1}, s_{j+1})$
\IF{$t < T_{discovery}$}
\STATE $\forall o_i \in \mathscr{O}, r_j \leftarrow f_i(s') - f_i(s)$
\STATE Train each option using its intrinsic reward (Eq. \ref{ddqn})
\STATE Minimize the generalized Laplacian (Eq. \ref{gl})
\ELSE
\STATE $\epsilon \leftarrow \text{LinearDecay}(\epsilon)$
\STATE Train main learner on extrinsic reward (Eq. \ref{ddqn})
\ENDIF
\STATE $s \leftarrow s'$
\ENDFOR
\end{algorithmic}
\label{algo:two_phase_dceo_option_discovery}
\end{algorithm}

\end{document}